\documentclass[twoside,11pt]{article}
\def\fighome {figures}%\structhome/
\usepackage{jmlr2e}
\usepackage[cmex10]{amsmath}
\usepackage{epsfig,epsf,psfrag,amssymb,amsfonts,latexsym,slashbox,graphicx,bm,cite,xcolor,url,array}
\usepackage[caption=false]{subfig}%,font=footnotesize
\usepackage{fixltx2e}
 \usepackage{cases}%[subnum]
\usepackage{verbatim}
\usepackage[mathscr]{eucal}
\usepackage{algorithm, algorithmic}
\usepackage[fleqn,tbtags]{mathtools}

     \DeclareMathOperator{\Norm}{norm}
 \DeclareMathOperator{\lbp}{LBP}

 \DeclareMathOperator{\Deg}{Deg}

\def\tha{{\mbox{\tiny th}}}
\DeclareMathOperator{\watts}{Watts}
\DeclareMathOperator{\girth}{Girth}

\DeclareMathOperator{\ER}{ER}
\DeclareMathOperator{\reg}{Reg}

\DeclareMathOperator{\hSigma}{\widehat{\Sigma}}

\DeclarePairedDelimiter\norm{\lVert}{\rVert}

\def\rinfnorm{\rVert_{\infty}}
\DeclarePairedDelimiter\infnorm{\lVert}{\rinfnorm}

 \DeclarePairedDelimiter\abs{\lvert}{\rvert}

 \def\0{{\bf 0}}

\DeclareMathOperator{\lea}{\overset{(a)}{\leq}}
\DeclareMathOperator{\leb}{\overset{(b)}{\leq}}
\DeclareMathOperator{\lec}{\overset{(c)}{\leq}}
\DeclareMathOperator{\led}{\overset{(d)}{\leq}}

\DeclareMathOperator{\eqa}{\overset{(a)}{=}}
\DeclareMathOperator{\eqb}{\overset{(b)}{=}}

\DeclareMathOperator{\gea}{\overset{(a)}{\geq}}
\DeclareMathOperator{\geb}{\overset{(b)}{\geq}}
\DeclareMathOperator{\gec}{\overset{(c)}{\geq}}

\def\eg{{e.g.,\ \/}}

\def\nn{\nonumber}

\def\qed{\hfill\hbox{${\vcenter{\vbox{
    \hrule height 0.4pt\hbox{\vrule width 0.4pt height 6pt
    \kern5pt\vrule width 0.4pt}\hrule height 0.4pt}}}$}}

\definecolor{myred}{rgb}{0.3,0.0,0.7}
\definecolor{dkg}{rgb}{0.1,0.7,0.2}
\definecolor{dkb}{rgb}{0.0,0.2,0.8}

\newcommand{\Gmsc}{\mathscr{G}}

 \def\hG{\widehat{G}}
 
 \def\hI{\widehat{I}}

 \def\hX{\widehat{X}}
\def\bfh{{\mathbf h}}

\def\bfw{{\mathbf w}}
\def\bfx{{\mathbf x}}

\def\bfA{{\mathbf A}}
\def\bfB{{\mathbf B}}

\def\bfD{{\mathbf D}}
\def\bfE{{\mathbf E}}
\def\bfF{{\mathbf F}}

\def\bfI{{\mathbf I}}
\def\bfJ{{\mathbf J}}

\def\bfR{{\mathbf R}}

\def\bfX{{\mathbf X}}

\def\mubf{\hbox{\boldmath$\mu$\unboldmath}}

\def\Sigmabf{\hbox{$\bf \Sigma$}}

\def\Qc{{\cal Q}}

\def\Sc{{\cal S}}

\def\Ebb{{\mathbb E}}
\def\Fbb{{\mathbb F}}

\def\Nbb{{\mathbb N}}

\def\Pbb{{\mathbb P}}

\def\tilJbf{{\widetilde{\bfJ}}}

\def\tilJbf{{\widetilde{\bfJ}}}

\newcommand{\bprfof}{\begin{proof_of}}
\newcommand{\eprfof}{\end{proof_of}}
\newcommand{\bprf}{\begin{myproof}}
\newcommand{\eprf}{\end{myproof}}
\newcommand{\bp}{\begin{psfrags}}
\newcommand{\ep}{\end{psfrags}}
\newcommand{\bl}{\begin{lemma}}
\newcommand{\el}{\end{lemma}}
\newcommand{\bt}{\begin{theorem}}
\newcommand{\et}{\end{theorem}}
\newcommand{\bc}{\begin{center}}
\newcommand{\ec}{\end{center}}
\newcommand{\bi}{\begin{itemize}}
\newcommand{\ei}{\end{itemize}}
\newcommand{\ben}{\begin{enumerate}}
\newcommand{\een}{\end{enumerate}}
\newcommand{\bd}{\begin{definition}}
\newcommand{\ed}{\end{definition}}
\def\beq{\begin{equation}}
\def\eeq{\end{equation}\noindent}
\def\beqn{\begin{eqnarray}}
\def\eeqn{\end{eqnarray} \noindent}
\def\beqnn{  \begin{eqnarray*}}
\def\eeqnn{\end{eqnarray*}  \noindent}
\def\bcase{  \begin{numcases}}
\def\ecase{\end{numcases}   \noindent}
\def\bsbcase{  \begin{subnumcases}}
\def\esbcase{\end{subnumcases}   \noindent}

\newenvironment{myproof}{\noindent{\em Proof:} \hspace*{1em}}{
    \hspace*{\fill} $\Box$ }
\newenvironment{proof_of}[1]{\noindent {\em Proof of #1: }}{\hspace*{\fill} $\Box$ }

\newcommand{\matplottc}[1]{               % single matlab plot twocolumn
        \unitlength .45truein
        \begin{center}
        \includegraphics{#1.ps}
        \end{picture}
        \end{center}
}

\def\psfancypar#1#2{\begingroup\def\par{\endgraf\endgroup\lineskiplimit=0pt}
               \setbox2=\hbox{\large\sc #2}
               \newdimen\tmpht \tmpht \ht2 \advance\tmpht by \baselineskip
               \font\hhuge=Times-Bold at \tmpht
               \setbox1=\hbox{{\hhuge #1}}
               \count7=\tmpht \count8=\ht1
               \divide\count8 by 1000 \divide\count7 by \count8
               \tmpht=.001\tmpht\multiply\tmpht by \count7
               \font\hhuge=Times-Bold at \tmpht
               \setbox1=\hbox{{\hhuge #1}}
               \noindent
                \hangindent1.05\wd1
               \hangafter=-2 {\hskip-\hangindent
               \lower1\ht1\hbox{\raise1.0\ht2\copy1}%
                \kern-0\wd1}\copy2\lineskiplimit=-1000pt}

\def\Kout{\setbox1=\hbox{\Huge\bf K}\hbox to
1.05\wd1{\hspace{.05\wd1}
}}
\def\Sout{\setbox1=\hbox{\Huge\bf S}\hbox to 1.05\wd1{\hspace{.05\wd1}
}}

\def\nbd{\mathcal{N}}

\def\Tsawi{{T_{\mathrm{saw}}(i;G)}}

\def\R{{\mathbb{R}}}

\allowdisplaybreaks[4]

\newcommand{\calX}{\mathcal{X}}

\newcommand{\calF}{\mathcal{F}}

\newcommand{\bX}{\mathbf{X}}

\newcommand{\bJ}{\mathbf{J}}

\newcommand{\calS}{\mathcal{S}}

\newcommand{\calN}{\mathcal{N}}

\newcommand{\calT}{\mathcal{T}}

\newcommand{\bx}{\mathbf{x}}

\newcommand{\bR}{\mathbb{R}}

\newcommand{\calU}{\mathcal{U}}

\newcommand{\bSigma}{\bm{\Sigma}}

\newcommand{\bI}{\mathbf{I}}

\newcommand{\bzero}{\mathbf{0}}

\newcommand{\hrho}{\widehat{\rho}}

\newcommand{\poly}{\mathrm{poly}}

\newcommand{\tilJ}{\widetilde{J}}

\newcommand{\threscondalgo}{\mathsf{CMIT}}

\renewcommand{\threscondalgo}{\mathsf{CCT}}
\newcommand{\thresmialgo}{\mathsf{CMIT}}
\def\Pep{P_e^{(p)}}
\def\frakG{\mathfrak{G}}
\def\frakE{\mathfrak{E}}
\newcommand{\Hb}{H_{\mathrm{b}}}
\def\fndot{\, \cdot \,}

\jmlrheading{}{}{}{6/11}{}{Animashree Anandkumar, Vincent Tan, and Alan Willsky}

\ShortHeadings{High-Dimensional Gaussian Graphical Model Selection}{Anandkumar, Tan, and Willsky}
\begin{document}
 \title{High-Dimensional Gaussian Graphical Model Selection:\\ Walk Summability and Local Separation Criterion}%under Correlation Decay
 %
%\author{Animashree Anandkumar\footnote{A. Anandkumar is with the Center for Pervasive Communications and Computing, Electrical Engineering and Computer Science Dept., University of California, Irvine, USA. Email: a.anandkumar@uci.edu},\,\, Vincent Y.F. Tan\footnote{V.Y.F. Tan is with the Electrical and Computer Engineering Dept., University of Wisconsin, Madison, USA. Email:vtan@wisc.edu} and Alan S. Willsky\footnote{Alan S. Willsky is with the Electrical Engineering and Computer Science Dept., Massachusetts Institute of Technology, Cambridge, USA. Email:willsky@mit.edu}}

\author{\name Animashree Anandkumar \email a.anandkumar@uci.edu \\
\addr Center for Pervasive Communications and Computing\\
Electrical Engineering and Computer Science\\
University of California, Irvine\\
Irvine, CA 92697\AND
\name Vincent Y. F. Tan \email    vtan@wisc.edu \\
\addr Department of Electrical and Computer Engineering\\
 University of Wisconsin-Madison\\
 Madison, WI 53706 \AND
 \name Alan S. Willsky \email willsky@mit.edu \\
\addr Stochastic Systems Group\\
Laboratory for Information and Decision Systems\\
Massachusetts Institute of Technology\\
Cambridge, MA 02139}

\editor{Martin Wainwright}

\maketitle

\begin{abstract}We consider the problem of high-dimensional Gaussian graphical model selection. We  identify a set of graphs for which an efficient estimation algorithm exists, and this algorithm is  based on thresholding of  empirical conditional covariances. Under a set of transparent conditions, we establish structural consistency (or {\em sparsistency}) for the proposed algorithm, when the number of samples $n=\Omega(J_{\min}^{-2} \log p)$, where $p$ is the number of variables and $J_{\min}$ is the minimum (absolute) edge potential of the graphical model. The sufficient conditions for sparsistency are based on the notion of {\em walk-summability} of the model and the presence of sparse {\em local vertex separators} in the underlying graph. We also derive novel non-asymptotic  necessary conditions on the number of samples required  for sparsistency. \end{abstract}


\section{Introduction}

%Such models have found widespread applications in a variety of areas including    computer vision, bio-informatics, financial modeling and social networks. For instance, graphical models have been  employed for contextual object recognition to improve detection performance based on object co-occurrences~\citep{choi_cvpr10} and   for
%financial modeling~\citep{Choi&etal:10JMLR}.

%There are mainly two tasks involving graphical models, viz., inference and learning. While inference has been extensively studied under various algorithms such as belief propagation~\cite{Wainwright&Jordan:08NOW}, high-dimensional learning of graphical models has been more recent. Learning of graphical models mainly falls into two categories: structure and parameter learning.

Probabilistic graphical models offer a powerful formalism for  representing  high-dimensional distributions succinctly. In an undirected  graphical model, the conditional independence relationships among the variables are represented in the form of an undirected graph.
Learning graphical models   using its observed  samples is an important task, and involves  both  structure and parameter estimation. While there are many techniques for parameter estimation (e.g., expectation maximization), structure estimation is arguably  more challenging. High-dimensional structure estimation is NP-hard for general models~\citep{Karger&Srebro:01SODA,Bogdanov&etal:Rand} and moreover, the number of samples available for learning is typically much smaller than the number of dimensions (or variables).

The complexity of structure estimation  depends crucially on the underlying graph structure.~\citet{Chow&Liu:68IT} established that structure estimation in tree models reduces to a maximum weight spanning tree problem and is thus  computationally efficient. However, a general characterization of graph families for which structure estimation is tractable  has so far been lacking. In this paper, we present such a characterization based on the so-called {\em local separation} property in graphs. It turns out that  a wide variety of (random) graphs satisfy this property (with probability tending to one) including large girth graphs, the Erd\H{o}s-R\'{e}nyi random graphs~\citep{Bollobas:book} and the power-law graphs~\citep{Chung:book}, as well as graphs with short cycles such as   the small-world graphs~\citep{watts1998collective} and other hybrid/augmented graphs~\citep[Ch. 12]{Chung:book}.

Successful structure estimation also relies on  certain assumptions on the parameters of the model, and these assumptions are tied to the specific algorithm employed. For instance, for convex-relaxation approaches~\citep{Mei06,Ravikumar&etal:08Arxiv},
the assumptions are based on certain {\em incoherence} conditions on the model, which are hard to interpret  as well as verify in general. In this paper, we present a set of transparent conditions for Gaussian graphical model selection based on {\em walk-sum} analysis~\citep{Malioutov&etal:06JMRL}.  Walk-sum analysis has been previously employed to analyze the performance of loopy belief propagation (LBP) and its variants in Gaussian graphical models. In this paper, we demonstrate that walk-summability also turns out to be a natural criterion for efficient structure estimation, thereby reinforcing its importance in characterizing  the tractability of   Gaussian graphical models.

\subsection{Summary of Results}

Our main contributions in this work are threefold. We propose a simple local algorithm for Gaussian graphical model selection, termed as conditional covariance threshold test ($\threscondalgo$)  based on a set of conditional covariance thresholding tests. Second, we derive sample complexity  results for our algorithm to achieve structural consistency (or sparsistency).  Third, we prove a novel non-asymptotic lower bound on  the sample complexity required by any learning algorithm to succeed.  We now elaborate on these contributions.

%The algorithm for model selection is a local algorithm which determines whether each node pair is an edge, based on a threshold on its empirical conditional covariance. Given a node pair, the algorithm computes the minimum empirical conditional covariance, over potential separators of cardinality bounded by a parameter $\eta\in \Nbb$. It is easy to see that the computational complexity of the algorithm is $O(p^{\eta+2})$, which is efficient when $\eta$ is small.

%\subsection{Conditional Covariance Threshold Test}
Our structure learning procedure is known as the Conditional Covariance Test\footnote{An analogous test is employed for Ising model selection in~\citep{AnandkumarTanWillsky:Ising11} based on conditional mutual information. We later note that conditional mutual information test has slightly worse sample complexity for learning Gaussian models.} $(\threscondalgo)$ and is outlined in Algorithm~\ref{algo:corr_thres_cond}.
Let $\threscondalgo(\bfx^n;\xi_{n,p},\eta)$ be the  output edge set from $\threscondalgo$ given $n$ i.i.d.\ samples $\bfx^n$, a threshold $\xi_{n,p}$ (that depends on both $p$ and $n$)  and a constant $\eta \in \Nbb$, which is related to the local vertex separation property (described later). The conditional covariance test  proceeds in the following manner.  First, the empirical absolute conditional covariances\footnote{Alternatively, conditional independence can be tested via sample partial correlations which can be computed via regression or recursion. See~\citep{kalisch2007estimating} for details.} are computed as follows: \[ \hSigma(i,j|S) := \hSigma(i,j)- \hSigma(i, S) \hSigma^{-1}(S,S)\hSigma(S,j), \] where $\hSigma(\cdot,\cdot)$ are the respective empirical variances. Note that $\hSigma^{-1}(S,S)$ exists when   the number of samples satisfies $n> |S|$ (which is the regime under consideration). The conditional covariance is thus computed for each node pair $(i,j)\in V^2$
 and   the conditioning set which achieves the minimum is found, over all subsets of cardinality at most $\eta$; if the minimum value exceeds the threshold $\xi_{n,p}$, then the node pair is declared as an edge.  See Algorithm~\ref{algo:corr_thres_cond}  for details.

The computational complexity of the algorithm is $O(p^{\eta+2})$,   which is efficient for small $\eta$.  For the so-called {\em walk-summable} Gaussian graphical models, the parameter $\eta$ can be interpreted as  an upper bound on the size of local vertex separators in the underlying graph. Many graph families have small $\eta$ and as such, are amenable to computationally efficient structure estimation by our algorithm. These include Erd\H{o}s-R\'{e}nyi random graphs, power-law graphs and small-world graphs, as discussed previously.

\begin{algorithm}[t]\begin{algorithmic}
\STATE Initialize $\hG^n_p= (V, \emptyset)$.\STATE
For each $i,j\in V$, if
\beq\min_{\substack{S\subset V\setminus\{i,j\}\\ |S|\leq \eta}}\abs{\hSigma(i,j| S)}> \xi_{n,p},\label{eqn:optsep}\eeq then  add $(i,j)$ to $\widehat{G}^n_p$.\\ \STATE
Output: $\widehat{G}^n_p$.\end{algorithmic}
\caption{Algorithm $\threscondalgo(\bfx^n;\xi_{n,p},\eta)$
for structure learning   using samples  $\bfx^n$.} \label{algo:corr_thres_cond}
\end{algorithm}

We establish that the proposed algorithm has a sample complexity of $n=\Omega(J_{\min}^{-2} \log p)$, where $p$ is the number of nodes  (variables)  and $J_{\min}$ is the minimum (absolute) edge potential in the model. As expected, the sample complexity improves when $J_{\min}$ is large, i.e., the model has strong edge potentials.  However, as we shall see,   $J_{\min}$ cannot be arbitrarily large for the model to be walk-summable. We derive the  minimum sample complexity for various graph families  and show that this  minimum is  attained when $J_{\min}$ takes the maximum possible value.

We also develop novel techniques to obtain necessary conditions for consistent structure estimation    of   Erd\H{o}s-R\'{e}nyi  random graphs and other ensembles with non-uniform distribution of graphs. We obtain non-asymptotic bounds on the number of samples $n$ in terms of the expected degree and the number of nodes of the model.  The techniques employed are information-theoretic in nature~\citep{Cover&Thomas:book}.   We cast the learning problem as a source-coding problem and develop necessary conditions which  combine the use of Fano's inequality with the so-called asymptotic equipartition property.

Our sufficient conditions for structural consistency are  based on walk-summability.  This characterization is novel to the best of our knowledge. Previously, walk-summable models have been extensively studied in the context of inference in Gaussian graphical models. As a by-product of our analysis, we also establish the correctness of loopy belief propagation for walk-summable Gaussian graphical models Markov on locally tree-like graphs (see Section~\ref{sec:lbp} for details). This suggests that walk-summability is a fundamental criterion for tractable learning and inference in Gaussian graphical models.

\subsection{Related Work}

%\subsubsection{Graphical Model Selection}

Given that structure learning of general graphical models is NP-hard~\citep{Karger&Srebro:01SODA,Bogdanov&etal:Rand}, the focus has been on characterizing classes of models on which learning is tractable. The seminal work of~\citet{Chow&Liu:68IT} provided an efficient implementation of maximum-likelihood  structure estimation for tree models  via a maximum weighted spanning tree algorithm. Error-exponent analysis of the Chow-Liu algorithm was studied \citep{Tan&etal:09ITsub,Tan&etal:10SP}  and  extensions to general forest models were  considered by \citet{Tan&etal:10JMLR} and \citet{Liu&etal:10forest}. Learning trees  with latent (hidden) variables \citep{Choi&etal:10JMLR} have also been studied recently.

For graphical models Markov on general graphs, alternative approaches are required for structure estimation. A recent paradigm for structure estimation is based on convex relaxation, where an estimate is obtained via convex optimization which incorporates  an $\ell_1$-based penalty term to encourage sparsity. For Gaussian graphical models, such approaches have been considered in~\citet{Mei06,Ravikumar&etal:08Arxiv,ABE:08}, and the sample complexity of the proposed algorithms have been analyzed.
A major disadvantage in using convex-relaxation methods is that the incoherence conditions required for consistent estimation are hard to interpret and it is not straightforward to characterize the class of models satisfying these conditions.

An alternative  to the convex-relaxation approach is the use of simple greedy local algorithms for structure learning. The conditions required for consistent estimation are typically more transparent, albeit somewhat restrictive.
~\citet{Bresler&etal:Rand} propose an algorithm for structure learning of general  graphical models Markov on    bounded-degree graphs, based on a series of conditional-independence tests.  \citet{Abbeel2006learning} propose an algorithm, similar in spirit, for learning factor graphs with bounded degree.
\citet{spirtes1995learning}, \citet{cheng2002learning}, \citet{kalisch2007estimating} and \citet{xie2008recursive} propose conditional-independence tests for learning Bayesian networks on directed acyclic graphs (DAG).
\citet{Sanghavi&etal:Allerton10} proposed a faster    greedy   algorithm, based on conditional entropy, for graphs with large girth and bounded degree.  However, all the  works~\citep{Bresler&etal:Rand,Abbeel2006learning,spirtes1995learning,
cheng2002learning,Sanghavi&etal:Allerton10}
require the maximum degree in the graph to be bounded ($\Delta=O(1)$) which is restrictive. We allow for graphs where the maximum degree can grow with the number of nodes. Moreover, we establish a natural tradeoff between the maximum degree and other parameters of the graph (e.g., girth) required for consistent structure estimation.

Necessary conditions for consistent graphical model selection   provide a lower bound on sample complexity  and have been explored before by \citet{Santhanam&Wainwright:08ISIT,Wang&etal:10ISIT}. These works consider    graphs drawn uniformly from the class of bounded degree graphs and establish that   $n=\Omega(\Delta^k \log p)$ samples are required for consistent structure estimation, in an $p$-node graph with maximum degree $\Delta$, where $k$ is typically a small positive integer.  However, a direct application of these methods yield poor lower bounds if the ensemble of graphs has a highly non-uniform distribution. This is the case with the ensemble of Erd\H{o}s-R\'{e}nyi random graphs~\citep{Bollobas:book}. Necessary conditions for structure estimation of Erd\H{o}s-R\'{e}nyi random graphs were derived for Ising models by~\citet{AnandkumarTanWillsky:10Stat} based on an information-theoretic covering argument. However, this approach is not directly applicable to the Gaussian setting. We present a novel approach for obtaining necessary conditions for Gaussian graphical model selection based on the notion of {\em typicality}. We characterize the set of typical graphs for the Erd\H{o}s-R\'{e}nyi ensemble and derive a modified form of Fano's inequality and obtain a non-asymptotic lower bound on sample complexity involving the average degree and the number of nodes.

We briefly also point to a large body of work on high-dimensional covariance selection under different notions of sparsity. Note that the assumption of a Gaussian graphical model Markov on a sparse graph is one such formulation. Other notions of sparsity include Gaussian models with sparse covariance matrices, or having a banded Cholesky factorization. Also, note that many works consider covariance estimation instead of selection and in general, estimation guarantees can be obtained under less stringent conditions. See~\citet{lam2009sparsistency}, \citet{rothman2008sparse}, \citet{huang2006covariance} and \citet{Bickel&Levina:08Stat} for details.

\paragraph{Paper Outline}

The paper is organized as follows. We introduce the system model in Section~\ref{sec:prelim}. We prove the main result of our paper regarding the structural consistency of conditional covariance thresholding test in Section~\ref{sec:method}. We prove necessary conditions for model selection in Section~\ref{sec:necessary}. In Section~\ref{sec:lbp}, we analyze the performance of loopy belief propagation in Gaussian graphical models. Section~\ref{sec:conclusion} concludes the paper. Proofs and additional discussion are provided in the appendix.

\section{Preliminaries and System Model}\label{sec:prelim}

\subsection{Gaussian Graphical Models}

A Gaussian graphical model is a family of jointly Gaussian distributions which factor in accordance to a given  graph.
Given a graph $G=(V,E)$, with $V  = \{1,\ldots, p\}$, consider a vector of Gaussian random   variables $\bfX=[X_1, X_2, \ldots, X_p]^T$, where each node $i \in V$    is associated with a scalar Gaussian random variable $X_i$. A Gaussian graphical model Markov on $G$ has a   probability density function (pdf) that may be  parameterized as
\beq\label{eqn:gauss}f_{\bfX}(\bfx) \propto \exp\left[-\frac{1}{2} \bfx^T \bfJ_G \bfx + \bfh^T \bfx\right],\eeq  where $\bfJ_G$ is a positive-definite symmetric matrix   whose sparsity pattern corresponds to that of the graph $G$. More precisely,   \beq \label{eqn:sparsity} J_G(i,j)=0 \iff (i,j) \notin G.\eeq The matrix $\bfJ_G$ is known as  the   potential or information matrix, the non-zero entries $J(i,j)$ as the edge potentials, and the vector $\bfh$ as the potential vector. A model is said to be {\em attractive} if $J_{i,j}\leq 0$ for all $i\neq j$. The form of parameterization in \eqref{eqn:gauss} is known as the information form and   is related to the standard mean-covariance parameterization of the Gaussian distribution as
\[  \mubf = \bfJ^{-1} \bfh,\quad \Sigmabf=\bfJ^{-1},\] where $\mubf:=\Ebb[\bfX]$ is the mean vector and  $\Sigmabf:=\Ebb[(\bfX-\mubf)(\bfX-\mubf)^T]$ is the covariance matrix.

We say that a jointly Gaussian random vector  $\bX$  with joint pdf $ f(\bx)$ satisfies local Markov property with respect to a graph $G$ if
\begin{equation}
f(x_i|\bx_{\calN(i)}) = f(x_i|\bx_{V\setminus i})
\end{equation}
holds for all nodes $i \in V$, where $\calN(i)$ denotes the set of neighbors of node $i\in V$ and,  $V\setminus i$ denotes the set of all nodes   excluding $i$. More generally, we say that  $\bfX$  satisfies the global Markov property, if for all disjoint sets $A,B\subset V$, we have
\beq f(\bx_A, \bx_B|\bx_S) = f(\bx_A|\bx_S) f(\bx_B|\bx_S).\eeq where    set $S$ is a {\em separator}\footnote{A set $S\subset V$ is a separator for sets $A$ and $B$ if the removal of nodes in $S$ partitions  $A$ and $B$ into distinct components.} of $A$ and $B$ The local and global Markov properties are equivalent for non-degenerate Gaussian distributions~\citep{Lauritzen:book}.

Our results  on structure learning depend on  the precision matrix $\bfJ$. Let \beq J_{\min} := \min_{(i, j)\in G} |J(i,j)|,\,\, J_{\max} :=\max_{(i, j)\in G} |J(i,j)|, \,\, D_{\min}:= \min_{i} J(i,i).\eeq
Intuitively, models with edge potentials which are   ``too small'' or ``too large'' are harder to learn   than those with comparable potentials. Since we consider the high-dimensional case where the number of variables $p$ grows, we allow the bounds $J_{\min}$, $J_{\max}$, and $D_{\min}$   to potentially scale with $p$.

%(which can, in turn,  depend on the number of nodes $p$). Note that previous algorithms~\cite{Mei06,Ravikumar&etal:08Arxiv} and lower bounds~\cite{Wang&etal:10ISIT} for Gaussian graphical models   all incorporate $J_{\min}$ in their results. We will also see that $J_{\min}$ and $J_{\max}$ need to scale in a certain manner in our setting to satisfy the so-called  {\em walk-summability} condition, provided  in \eqref{eqn:walksummable},  and is   discussed in Section~\ref{sec:walkrandom}.

The {\em partial correlation coefficient} between variables $X_i$ and $X_j$, for $i \neq j$, measures their conditional covariance given all other variables. These are computed by normalizing the off-diagonal values of the information matrix, i.e.,
\beq R(i,j) := \frac{\Sigma(i,j| {V\setminus\{i,j\}})
}{\sqrt{\Sigma(i,i| {V\setminus\{i,j\}})\Sigma(j,j| {V\setminus\{i,j\}})}}
= - \frac{J(i,j)}{\sqrt{J(i,i) J(j,j)}}.\label{eqn:parcor}\eeq For all $i \in V$, set $R(i,i)=0$. We henceforth refer to $\bfR$ as the partial correlation matrix.

An important sub-class of   Gaussian graphical models  of the form in \eqref{eqn:potential_normalized} are the  {\em walk-summable} models~\citep{Malioutov&etal:06JMRL}. A Gaussian model is said to be $\alpha$-walk summable if
\beq  \norm{\overline{\bfR}}\leq \alpha<1,\eeq
where  $\overline{\bfR}:=[|R(i,j)|]$  denotes the entry-wise absolute value of the partial correlation matrix $\bfR$   and $\norm{\fndot}$ denotes the spectral or 2-norm of the matrix, which for symmetric matrices, is given by the maximum absolute eigenvalue.

In other words, walk-summability   means that an  attractive model formed by taking the absolute values of the partial correlation matrix of the Gaussian graphical model is also valid (i.e., the corresponding potential matrix is positive definite). This immediately implies that attractive models  form  a sub-class of walk-summable models.  For detailed discussion on walk-summability, see Section~\ref{sec:backgroundwalk}.

\subsection{Tractable Graph Families}

We consider the class of Gaussian  graphical models Markov on a  graph $G_p$  belonging to some ensemble $\Gmsc(p)$ of graphs with $p$ nodes. We consider the high-dimensional learning  regime, where both  $p$ and the number of samples $n$ grow  simultaneously; typically, the growth of $p$ is much faster than that of $n$. We emphasize that in our formulation the graph ensemble $\Gmsc(p)$ can either be deterministic or random -- in the latter, we also specify a probability measure over the set of graphs in $\Gmsc(p)$. In the setting where $\Gmsc(p)$ is a random-graph ensemble, let $P_{\bfX, G}$ denote the joint probability distribution of the variables $\bfX$   and the graph $G\sim\Gmsc(p)$, and let $f_{\bfX|G}$ denote the conditional (Gaussian) density of the variables Markov on the  given   graph $G$. Let $P_G$ denote the probability distribution of graph $G$  drawn from a random ensemble $\Gmsc(p)$. We use the term  {\em almost every} (a.e.) graph $G$ satisfies a certain property $\Qc$ if \[ \lim_{p\to \infty} P_G[G\mbox{ satisfies }\Qc]=1.\] In other words, the property $\Qc$ holds asymptotically almost surely\footnote{Note that the term a.a.s.  does not apply to deterministic graph ensembles $\Gmsc(p)$ where   no randomness  is assumed, and in this setting, we assume that the property $\Qc$   holds for every graph in the ensemble.}  (a.a.s.) with respect to the random-graph ensemble $\Gmsc(p)$. Our conditions and theoretical  guarantees will be based on this notion for random graph ensembles. Intuitively, this means that  graphs that have a vanishing probability of occurrence as $p \to \infty$ are ignored.

\begin{figure}[t]
\centering
\bp\psfrag{j}[l]{$j$}\psfrag{k1}[l]{$a$}\psfrag{k2}[l]{$b$}
\psfrag{k3}[l]{$c$}\psfrag{k4}[l]{$d$}\psfrag{A}[l]{$A$}
\psfrag{U}[l]{$\calU(j;\Tsawi)$}\psfrag{tilU}[l]{$\widetilde{\calU}(j,a;\Tsawi)$}
\psfrag{i}[l]{$i$}\psfrag{S}[l]{$\Sc(i,j)$}
\includegraphics[width=1.8in,height=1.8in]{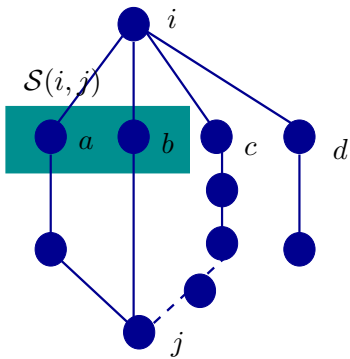}
\ep
\caption{Illustration of $l$-local separator set $\calS(i,j;G,l)$ for the graph shown above with  $l = 4$. Note that $\nbd(i)=\{a,b,c,d\}$ is the neighborhood of $i$  and the $l$-local separator set $\calS(i,j;G,l)=\{a,b\}\subset \nbd(i;G)$. This is because  the path along $c$ connecting $i$ and $j$ has a length greater than $l$ and hence node $c\notin \Sc(i,j;G,l)$.  }
\label{fig:pseparator}
\end{figure}

We now characterize the ensemble of graphs amenable for consistent structure estimation under our formulation. To this end, we define the concept of local separation in graphs. See Fig.~\ref{fig:pseparator} for an illustration. For $\gamma\in\Nbb$, let $B_\gamma(i;G )$ denote the set of vertices within distance $\gamma$ from $i$ with respect to graph $G$. Let  $H_{\gamma, i}:=G(B_\gamma(i))$ denote the subgraph of $G$ spanned by $B_\gamma(i;G)$, but in addition, we retain the nodes not in $B_\gamma(i)$ (and remove the corresponding edges).  Thus, the number of vertices in $H_{\gamma, i}$ is $p$.

\bd[$\gamma$-Local Separator]\label{def:sep} Given a graph $G$, a $\gamma$-{\em local separator} $S_\gamma(i,j)$ between $i$ and $j$,   for $(i,j)\notin G$,  is a {\em minimal} vertex separator\footnote{A minimal separator is a separator of smallest cardinality.} with respect to the subgraph $H_{\gamma,i} $. In addition,  the parameter $\gamma$ is referred to as the {\em path threshold} for  local separation.
\ed

In other words, the $\gamma$-local separator $S_\gamma(i,j)$ separates nodes $i$ and $j$ with respect to paths in $G$ of length at most $\gamma$.  We now characterize the ensemble of graphs based on the size of local separators.

%We  immediately conclude that a.e. graph $G\sim \Gmsc(p;\eta,\gamma)$ has $\gamma$-local separators $ $ for all $(i,j)\notin G$.

\bd[$(\eta,\gamma)$-Local Separation Property]\label{def:localpath}An ensemble of graphs satisfies $(\eta,\gamma)$-local separation  property  if for a.e. $G_p$ in the ensemble, \beq  \max_{(i,j)\notin G_p}|S_\gamma(i,j)|\leq \eta.\label{eqn:localpath}\eeq We denote such a graph ensemble by $\Gmsc(p;\eta,\gamma)$.\ed

In Section~\ref{sec:method}, we propose an efficient algorithm for graphical model selection when the underlying graph belongs to a  graph ensemble $\Gmsc(p;\eta, \gamma)$ with sparse local separators  (i.e., small $\eta$, for $\eta$ defined in \eqref{eqn:localpath}). We will see that the computational complexity of our proposed algorithm scales as $O(p^{\eta+2})$.  We now provide examples of several  graph families satisfying \eqref{eqn:localpath}.

%\subsubsection{Examples}\label{sec:graphdetails}

%Recall the Definition~\ref{def:localpath} on local-path property that a.e. graph $G$ drawn from the ensemble $\Gmsc(p;\eta,\gamma)$ has at most $\eta$ paths between any two nodes of length at most $\gamma$, for some $\eta,\gamma\in \Nbb$.

\subsubsection*{Example 1: Bounded-Degree}

We now show that the local-separation property holds for a rich class of graphs. Any   (deterministic or random)  ensemble   of degree-bounded graphs $\Gmsc_{\Deg}(p, \Delta)$ satisfies $(\eta,\gamma)$-local separation property with $\eta=\Delta$ and arbitrary $\gamma\in \Nbb$. If we do not impose any further constraints on $\Gmsc_{\Deg}$, the computational complexity of our proposed algorithm scales as $O(p^{\Delta+2})$ (see also~\citet{Bresler&etal:Rand}  where the computational complexity is comparable). Thus, when $\Delta$ is large, our proposed algorithm and the one in~\citet{Bresler&etal:Rand} are computationally intensive.   Our goal in this paper is to relax the usual bounded-degree  assumption and to consider   ensembles of graphs $\Gmsc(p)$ whose maximum degrees may grow with the number of nodes $p$.  To this end, we discuss other structural constraints which can lead to graphs with sparse local separators.

\subsubsection*{Example 2: Bounded Local Paths}\label{sec:localpaths}

Another sufficient condition\footnote{For any graph satisfying $(\eta,\gamma)$-local separation property, the number of vertex-disjoint paths of length at most $\gamma$ between any two non-neighbors is bounded above by $\eta$, by appealing to Menger's theorem for bounded path lengths~\citep{lovász1978mengerian}. However, in the definition of local-paths property, we consider all distinct paths of length at most $\gamma$ and not just vertex disjoint paths.} for the $(\eta,\gamma)$-local separation property in Definition~\ref{def:localpath} to hold  is that there are at most $\eta$ paths  of length at most $\gamma$ in $G$ between any two nodes (henceforth, termed as the $(\eta,\gamma)$-{\em local paths property}). In other words, there are at most $\eta-1$ number of overlapping\footnote{Two cycles are said to overlap if they have common vertices.} cycles of length smaller than $2\gamma$.

In particular,   a special case of the local-paths property described above  is the so-called   girth property. The {\em girth} of a graph is the length of the shortest cycle. Thus, a graph with girth $g$ satisfies $(\eta, \gamma)$-local separation property with $\eta=1$ and $\gamma=g$.  Let $\Gmsc_{\girth}(p;g)$ denote the ensemble of graphs with girth at most $g$. There are many graph constructions which lead to large girth.  For example,    the bipartite Ramanujan graph~\citep[p. 107]{Chung:book2} and  the random Cayley graphs~\citep{gamburd2009girth}   have large girths.

%For example,   the ensemble of $\Delta$-random regular graphs, denoted by $\Gmsc_{\reg}(p, \Delta)$, which  is the uniform ensemble of  regular graphs with  degree $\Delta$ has a girth of  $\Theta(\log_{\Delta-1} p)$~\citep[p. 107]{Chung:book2}. Other constructions such as the bipartite Ramanujan graph also have large girths~\citep[p. 107]{Chung:book2}.

The girth condition can be weakened to allow for a small number of short cycles, while not allowing for typical node neighborhoods  to contain short cycles. Such graphs are termed as {\em locally tree-like}. For instance, the ensemble of  Erd\H{o}s-R\'{e}nyi graphs $\Gmsc_{\ER}(p, c/p)$, where an edge between any node pair appears with a probability  $c/p$,  independent of  other node pairs, is locally tree-like.
The parameter $c$ may grow   with $p$, albeit at a controlled rate for tractable structure learning. We make this more precise in Example 3 in  Section~\ref{sec:assume}.  The proof of the following result may be found in~\citep[Lemma 3]{AnandkumarAvinatanKelner:11Arxiv}.

%The ensemble of random regular graphs $\Gmsc_{\reg}(p, \Delta)$  satisfies the $(\eta,\gamma)$-local separation  property   a.a.s. with $\eta=2$ and any $\gamma\in\Nbb$ that satisfies  \[ (\Delta-1)^{2\gamma-1}=o(p).\] The proofs for random regular graphs and ER graphs can be found in~\citep[(2.10)]{mckay2004short} and

\begin{proposition}[Random Graphs are Locally Tree-Like]\label{prop:ERlocaltree}
 The ensemble of Erd\H{o}s-R\'{e}nyi graphs $\Gmsc_{\ER}(p, c/p)$ satisfies the $(\eta,\gamma)$-local separation property in \eqref{eqn:localpath} with   \beq \eta=2, \,\, \gamma\leq \frac{\log p}{4\log c}.\label{eqn:gammaER}\eeq\end{proposition}

Thus, there are at most two paths  of length smaller than $\gamma$ between any two nodes in Erd\H{o}s-R\'{e}nyi graphs a.a.s, or equivalently,    there are no overlapping  cycles of length smaller than $2 \gamma$ a.a.s. Similar observations apply for the more general {\em scale-free} or {\em power-law} graphs~\citep{Chung:book,dommers2010ising}.
Along similar lines, the ensemble of $\Delta$-random regular graphs, denoted by $\Gmsc_{\reg}(p, \Delta)$, which  is the uniform ensemble of  regular graphs with  degree $\Delta$  has no overlapping cycles of length at most $\Theta(\log_{\Delta-1} p)$ a.a.s.~\citep[Lemma 1]{mckay2004short}.

\subsubsection*{Example 3: Small-World Graphs}

The previous two examples showed  local separation holds under two different conditions: bounded  maximum degree and bounded number of  local paths. The former class of graphs  can have short cycles but the maximum degree needs to be constant, while the latter class  of graphs  can have a large maximum degree  but  the number of overlapping short cycles needs to be small. We now provide instances  which incorporate both these features: large degrees and short cycles, and yet satisfy the local separation property.

The class of hybrid graphs or augmented graphs~\citep[Ch. 12]{Chung:book} consists of graphs which are  the  union of two graphs: a ``local'' graph having short cycles and a ``global'' graph having small average distances. Since the hybrid graph is  the union of these local and global graphs, it    has both large degrees and short cycles. The simplest model $\Gmsc_{\watts}(p, d, c/p)$, first studied by~\citet{watts1998collective}, consists of the union of a $d$-dimensional grid and an  Erd\H{o}s-R\'{e}nyi random graph with parameter $c$.
It is easily seen that a.e. graph   $G\sim\Gmsc_{\watts}(p, d, c/p)$ satisfies $(\eta,\gamma)$-local separation property in \eqref{eqn:localpath}, with \[ \eta=d+2, \,\,\gamma\leq\frac{\log p}{4\log c}.\]  Similar observations apply for more general hybrid graphs studied in~\citep[Ch. 12]{Chung:book}.

%Thus, when $d$ is a small constant and $c=O(\poly\log p)$, we show that  our proposed algorithm in Section~\ref{sec:method}  is structurally consistent in high-dimensions  and has a computational complexity of $O(p^{d+4})$.

\subsubsection*{Counter-example: Dense Graphs}
While the above examples illustrate that a large class of graphs satisfy the local separation criterion, there indeed exist graphs which do not satisfy it. Such graphs tend to be ``dense'', i.e., the number of edges scales super-linearly in the number of nodes. For instance, the Erd\H{o}s-R\'{e}nyi graphs $\Gmsc_{\ER}(p, c/p)$ in the dense regime, where the average degree scales as $c=  \Omega(p^2)$. In this regime, the node degrees as well as the number of short cycles grow with $p$  and thus, the size of the local separators also grows with $p$. Such graphs are hard instances for our algorithm.

\section{Guarantees for Conditional Covariance Thresholding}\label{sec:method}

\subsection{Assumptions}\label{sec:assume}

\begin{enumerate}
\item[(A1)] {\bf  Sample Scaling Requirements:} We consider the asymptotic setting where both the number of variables (nodes) $p$ and the number of samples $n$ tend to infinity. We assume that the parameters $(n,p, J_{\min})$ scale in the following fashion:\footnote{The notations $\omega(\cdot)$, $\Omega(\cdot)$ refer to asymptotics as the number of variables $p \to \infty$.}
\beq \label{eqn:sample_complexity2}n
 =\Omega(J_{\min}^{-2} \log p).\eeq   We require that the number of nodes $p \to \infty$ to exploit the local separation properties of the class of graphs under consideration.

\item[(A2)] {\bf $\alpha$-Walk-summability:} The Gaussian graphical model Markov  on $G_p\sim \Gmsc(p)$ is $\alpha$-walk summable a.a.s., i.e.,  \beq \label{eqn:walksummable} \norm{\overline{\bfR}_{G_p}}\leq \alpha<1,\quad \mbox{a.e. }G_p\sim \Gmsc(p),\eeq
where $\alpha$ is a constant (i.e.,  not a function of $p$),  $\overline{\bfR}:=[|R(i,j)|]$  is the entry-wise absolute value of the   partial correlation matrix $\bfR$ and $\norm{\cdot}$ denotes the spectral norm.

% To emphasize the role of $\alpha$ (which is not dependent on $p$), we say that the random ensemble $\Gmsc(p)$ is $\alpha$-walk-summable if \eqref{eqn:walksummable} is satisfied a.a.s.

\item[(A3)]{\bf Local-Separation Property: }  We assume that the ensemble of graphs $\Gmsc(p;\eta,\gamma)$ satisfies the $(\eta,\gamma)$-local separation property with   $\eta,\gamma$ satisfying:   \beq \eta=O(1),\,\,\,J_{\min} D^{-1}_{\min} \alpha^{-\gamma} = \omega(1),\label{eqn:gamma} \eeq where $\alpha$ is given by \eqref{eqn:walksummable} and $D_{\min}:=\min_i J(i,i)$ is the minimum diagonal entry of the potential matrix $\bfJ$.

\item[(A4)]{\bf Condition on Edge-Potentials: }The minimum   absolute edge potential of an $\alpha$-walk summable Gaussian graphical model satisfies  \beq D_{\min}(1-\alpha) \min_{(i,j)\in G_p} \frac{J(i,j)}{K(i,j)}>1+\delta,\label{eqn:minmax}\eeq
   for almost every $G_p\sim \Gmsc(p)$,  for some $\delta>0$ (not depending on $p$) and\footnote{Here and in the sequel, for $A,B\subset V$, we use the notation $\bfJ(A,B)$ to denote the sub-matrix of $\bfJ$ indexed by rows in $A$ and columns in $B$.}
\[K(i,j):= \norm{\bfJ(V\setminus \{i,j\}, \{i,j\})}^2 , \] is the spectral norm of the submatrix of the potential matrix $\bfJ$, and $D_{\min}:=\min_i J(i,i)$ is the minimum diagonal entry of $\bfJ$.  Intuitively,  \eqref{eqn:minmax} limits the extent of non-homogeneity in the model and the extent of overlap of neighborhoods.  Moreover, this assumption  is   not required for consistent graphical model selection when the model is  attractive ($J_{i,j}\leq 0$ for $i \neq j$).\footnote{The assumption (A5)   rules out the possibility that   the neighbors are marginally independent. See Section~\ref{sec:nbd} for details.}
% We also assume   a generic choice of edge potentials satisfying the bounds $J_{\min}$ and $J_{\max}$, i.e., except over a set of vanishing Lebesgue measure.
% Note that $J_{\min}$ and $J_{\max}$ may be functions of $p$. The relation in \eqref{eqn:minmax} stipulates how these parameters should scale for consistent structure recovery. Intuitively, the above condition limits the non-homogeneity in the model.

\item[(A5)]{\bf Choice of threshold $\xi_{n,p}$: }
The threshold $\xi_{n,p}$ for graph estimation under $\threscondalgo$ algorithm   is chosen as a function of  the number of nodes $p$, the number of samples $n$, and  the minimum edge potential $J_{\min}$ as follows:\beq  \xi_{n,p}= O(J_{\min}),\,\, \xi_{n,p} = \omega\left(\frac{\alpha^{\gamma}}{D_{\min}}\right),\,\,\xi_{n,p}= \Omega\left(\sqrt{\frac{\log p}{n}}\right),\label{eqn:xi}\eeq
where $\alpha$ is given by \eqref{eqn:walksummable}, $D_{\min}:=\min_i J(i,i)$ is the minimum diagonal entry of the potential matrix $\bfJ$, and $\gamma$ is the path-threshold \eqref{eqn:localpath} for the $(\eta,\gamma)$-local separation property to hold.

\end{enumerate}

Assumption (A1) stipulates how $n$, $p$ and $J_{\min}$ should scale for consistent graphical model selection, i.e.,    the sample complexity. The sample size $n$ needs to be sufficiently large with respect to the number of variables $p$ in the model for consistent structure reconstruction. Assumptions (A2) and (A4) impose constraints on the model parameters.   Assumption (A3) restricts the class of graphs under consideration. To the best of our knowledge, all   previous works dealing with graphical model selection,  \eg \citet{Mei06}, \citet{Ravikumar&etal:08Arxiv}, also impose some conditions for consistent graphical model selection. Assumption (A5) is with regard to the choice of a suitable threshold $\xi_{n,p}$ for thresholding conditional covariances.
 In the sequel,  we  compare the conditions for consistent recovery after presenting our main theorem.

\subsubsection*{Example 1: Degree-Bounded Ensembles}
To gain a better understanding of conditions     (A1)--(A5), consider  the ensemble of graphs  $\Gmsc_{\Deg}(p;\Delta)$ with bounded degree  $\Delta\in \Nbb$. It can be established that for the walk-summability condition in (A3) to hold,\footnote{We can provide improved bounds for random-graph ensembles. See Section~\ref{sec:walkrandom} for details.} we require that for normalized precision matrices  $(J(i,i)=1)$, \beq J_{\max}= O\left(\frac{1}{\Delta}\right).\eeq See Section~\ref{sec:walkrandom} for detailed discussion. When the minimum potential achieves the bound $(J_{\min}= \Theta (1/\Delta))$,  a sufficient condition for (A3) to hold is given by  \beq \label{eqn:condDelta} \Delta \alpha^\gamma = o(1),\eeq where $\gamma$ is the  path threshold   for the local-separation property to hold according to Definition~\ref{def:localpath}. Intuitively, we require a larger path threshold $\gamma$, as the degree bound $\Delta$ on the graph ensemble increases.

Note that \eqref{eqn:condDelta} allows for the degree bound $\Delta$ to grow with the number of nodes as long as the path threshold $\gamma$ also grows appropriately. For example,   if the maximum degree scales as $\Delta=O( \poly(\log p))$ and the path-threshold scales as $\gamma=O(\log \log p)$, then  \eqref{eqn:condDelta} is satisfied. This implies that graphs with fairly large degrees and short cycles can be recovered successfully using our algorithm.

\subsubsection*{Example 2: Girth-Bounded Ensembles}
The  condition in \eqref{eqn:condDelta} can be  specialized for the ensemble of girth-bounded graphs $\Gmsc_{\girth}(p;g)$ in a straightforward manner as
\beq \label{eqn:condgirth} \Delta \alpha^g = o(1),\eeq
where $g$ corresponds to the {\em girth} of the graphs in the ensemble. The condition in \eqref{eqn:condgirth} demonstrates a   natural  tradeoff between the girth and the maximum degree; graphs with large  degrees can be learned efficiently if their girths are large.
Indeed, in the extreme case  of trees which have infinite girth, in accordance with \eqref{eqn:condgirth}, there is no constraint on node degrees for successful recovery and recall that the Chow-Liu algorithm~\citep{Chow&Liu:68IT} is an efficient method for model selection on tree distributions.

\subsubsection*{Example 3: Erd\H{o}s-R\'{e}nyi and Small-World Ensembles}
We can also conclude that a.e.  Erd\H{o}s-R\'{e}nyi graph $G\sim \Gmsc_{\ER}(p, c/p)$     satisfies \eqref{eqn:gamma} when $c= O(\poly(\log p))$ under the best-possible scaling of $J_{\min}$ subject to the walk-summability constraint in \eqref{eqn:walksummable}.

This is because it can be shown that $J_{\min}=O(1/\sqrt{\Delta})$ for walk-summability in \eqref{eqn:walksummable} to hold. See Section~\ref{sec:walkrandom} for details. Noting that
a.a.s., the maximum degree $\Delta$ for $G\sim \Gmsc_{\ER}(p, c/p)$   satisfies \[\Delta  = O\left(\frac{\log p \log c }{ \log \log p}\right),\] from~\citep[Ex. 3.6]{Bollobas:book} and $\gamma=O(\frac{\log p}{\log c})$ from \eqref{eqn:gammaER}. Thus, the Erd\H{o}s-R\'{e}nyi graphs  are  amenable to successful
recovery when the average degree $c= O(\poly(\log p))$.
Similarly, for the small-world ensemble $\Gmsc_{\watts}(p, d, c/p)$, when $d=O(1)$ and $c= O(\poly(\log p))$, the graphs are amenable for consistent estimation.

%We now also comment on the condition on $J_{\min}, \alpha, \gamma$ in  \eqref{eqn:gamma} required for successful graphical model selection, where $\gamma$ is the threshold for the local-path property  $(\eta,\gamma)$ in the graph ensemble $\Gmsc(p;\eta,\gamma)$ under consideration. Recall that \eqref{eqn:gamma} entails that $J_{\min} \alpha^{-\gamma} = \omega(1)$.  An upper bound on $\gamma$ can be obtained  on graphs bounded by maximum degree $\Delta$ and under assumption (A4) in Section~\ref{sec:assume},  \[ J_{\min}= \Omega(J^2_{\max})= O\left(\frac{1}{\Delta^2}\right),\] and this implies that we require  \beq \label{eqn:condDeltal} \Delta^2 \alpha^\gamma = o(1).\eeq
%If we consider the class of girth-bounded graphs $(\eta=1, \gamma=g)$,  we require the above condition on the degree and the girth of the graph. Intuitively, we require a higher girth for graphs with larger degree. Indeed in the extreme case of trees, we have the classical result that trees of any degree can be recovered.

\subsection{Consistency of Conditional Covariance Thresholding}

Assuming (A1) -- (A5),  we  now state our main result. The proof of this result and the auxiliary lemmata for the proof can be found in Sections~\ref{proof:graphs} and Section~\ref{proof:sample}.

\bt[Structural consistency of $\threscondalgo$]\label{thm:corr_thres_cond}
For structure learning  of   Gaussian graphical  models Markov on a graph $G_p \sim \Gmsc(p;\eta,\gamma)$,   $\threscondalgo(\bfx^n;\xi_{n,p},\eta)$ is consistent for a.e. graph $G_p$. In other words,  \beq \lim_{\substack{n,p\to \infty\\ n=\Omega(J_{\min}^{-2}\log p) }}P\left[\threscondalgo \left(\{\bx^n\};\xi_{n,p},\eta \right) \neq G_p \right]=0 \eeq  \et

\vspace{1em}

%From \eqref{eqn:xi}, it is   evident that we require \[ J_{\min}=\omega\left(\sqrt{\frac{\log p}{n}}\right).\]

\noindent{\bf Remarks: }\ben
\item {\bf Consistency guarantee: }The $\threscondalgo$ algorithm consistently recovers the structure of Gaussian graphical models asymptotically, with probability tending to one, where the probability  measure is with respect to both the random graph (drawn from the  ensemble $\Gmsc(p;\eta,\gamma)$ and the samples (drawn from $\prod_{i=1}^n f(\bx_i|G)$).

%\item {\bf Examples of graphs amenable for successful recovery: }A large family of graphs satisfy the requirements discussed previously and are thus amenable for successful recovery, according to Theorem~\ref{thm:corr_thres_cond}. For details, see Section~\ref{sec:graphdetails}. For example, the requirement in \eqref{eqn:condDelta} is satisfied for girth-bounded graphs,  Erd\H{o}s-R\'{e}nyi random graphs $\Gmsc_{\ER}(p, c/p)$, when  $c= O(\poly(\log p))$, random regular graphs $\Gmsc_{\reg}(p, \Delta)$ when $\Delta=\poly(\log p)$, and  more generally, we can allow for maximum degree to scale as $\Delta=O( \poly(\log p))$ and for the short-path threshold to scale as $\gamma= O(\log \log p)$. This implies that graphs with fairly large degrees and short cycles can be recovered  successfully by our algorithm under walk-summability.

\item {\bf Analysis of sample complexity: }The above result states that the sample complexity  for the $\threscondalgo$ $(n =\Omega(J_{\min}^{-2} \log p))$, which improves when the minimum edge potential $J_{\min}$ is large.\footnote{Note that the sample complexity also implicitly depends on walk-summability parameter $\alpha$ through \eqref{eqn:gamma}.} This is intuitive since the edges have stronger potentials in this case. On the other hand, $J_{\min}$ cannot be arbitrarily large since the $\alpha$-walk-summability assumption in \eqref{eqn:walksummable}  imposes an upper bound on $J_{\min}$. The minimum  sample complexity  (over different parameter settings)  is attained  when $J_{\min}$ achieves this upper bound.   See Section~\ref{sec:walkrandom} for details. For example, for any degree-bounded  graph ensemble $\Gmsc(p,\Delta)$ with maximum degree $\Delta$, the minimum sample complexity is $n = \Omega(\Delta^2 \log p)$ i.e., when $J_{\min}=\Theta(1/\Delta)$, while for  Erd\H{o}s-R\'{e}nyi random graphs, the minimum sample complexity can be improved to $n =\Omega(\Delta \log p)$, i.e., when $J_{\min}=\Theta(1/\sqrt{\Delta})$.

\item {\bf Comparison with~\citet{Ravikumar&etal:08Arxiv}: }
The work by~\citet{Ravikumar&etal:08Arxiv} employs an $\ell_1$-penalized likelihood estimator  for structure estimation in Gaussian graphical models. Under the so-called incoherence conditions, the  sample complexity is $n=\Omega((\Delta^2+J_{\min}^{-2}) \log p)$.
Our sample complexity in \eqref{eqn:sample_complexity2} is the same in terms of    its  dependence on $J_{\min}$, and there is no explicit dependence on the maximum degree  $\Delta$.
Moreover, we have a transparent sufficient condition in terms of $\alpha$-walk-summability in \eqref{eqn:walksummable}, which directly imposes scaling conditions on $J_{\min}$.

%In contrast,  the precise scaling of parameters to achieve incoherence conditions in~\cite{Ravikumar&etal:08Arxiv} are not straightforward to verify. Also, our algorithm is a simple local test and can be efficiently implemented.

\item {\bf Comparison with~\citet{Mei06}: }The work by~\citet{Mei06} considers $\ell_1$-penalized linear regression for neighborhood selection of Gaussian graphical models and establish a  sample complexity of $n=\Omega((\Delta+J_{\min}^{-2}) \log p)$. We note that our guarantees allow for graphs which do not necessarily  satisfy the conditions imposed by~\citet{Mei06}. For instance,  the assumption of neighborhood stability (assumption 6 in~\citep{Mei06}) is hard to verify in general, and the relaxation of this assumption corresponds to the class of models with  diagonally-dominant covariance matrices. Note that the class of Gaussian graphical models with diagonally-dominant covariance matrices    forms a strict sub-class of walk-summable models, and thus satisfies assumption (A2)  for the  theorem to hold.  Thus, Theorem~\ref{thm:corr_thres_cond} applies to a larger class of Gaussian graphical models  compared to~\citet{Mei06}. Furthermore, the conditions for successful recovery in Theorem~\ref{thm:corr_thres_cond} are arguably more transparent.

%the  condition of small overlap of neighborhoods (assumption 4 in~\citet{Mei06})  is not satisfied asymptotically almost surely  by  Erd\H{o}s-R\'{e}nyi random graphs. Moreover,

\item {\bf Comparison with Ising models: }Our above result for learning Gaussian graphical models  is analogous to structure estimation of Ising models subject to an upper bound on the edge potentials~\citep{AnandkumarTanWillsky:Ising11}, and we characterize such a regime as a {\em conditional uniqueness} regime. Thus,   walk-summability is the analogous condition for Gaussian models.

    %The difference is that the sample complexity for Gaussian models in \eqref{eqn:sample_complexity2}, $n=\omega(J_{\min}^{-2} \log p)$ is worse than in the Ising model setting, which is $n=\omega(\log p)$. This is because under Ising model, we can have correlation decay with $J_{\min}=\Omega(1)$, while we require $J_{\min}$ to decay in \eqref{eqn:Rmin} to ensure walk-summability in Gaussian models.
\een

\paragraph{Proof Outline}  We first analyze the scenario when  exact statistics are available. (i) We establish that for any two non-neighbors $(i,j)\notin G$, the minimum conditional covariance in \eqref{eqn:optsep}   (based on exact statistics) does not exceed the threshold $\xi_{n,p}$. (ii) Similarly, we also  establish that the  conditional covariance in \eqref{eqn:optsep} exceeds the threshold $\xi_{n,p}$ for all neighbors $(i,j)\in G$. (iii) We then extend these results to empirical versions  using concentration bounds.

\subsubsection{Performance of Conditional Mutual Information Test}

We now employ the conditional mutual information test, analyzed in~\citet{AnandkumarTanWillsky:Ising11} for Ising models, and note that it has slightly worse sample complexity than using conditional covariances. Using the threshold $\xi_{n,p}$ defined in \eqref{eqn:xi}, the conditional mutual information test $\thresmialgo$ is given by the threshold test
\beq\min_{\substack{S\subset V\setminus\{i,j\}\\ |S|\leq \eta}}\hI(X_i;X_j| \bfX_S)> \xi^2_{n,p},\label{eqn:optsepnew}\eeq and node pairs   $(i,j)$ exceeding the threshold are added to the estimate $\widehat{G}^n_p$.
 Assuming (A1) -- (A5),  we have the following result.

\bt[Structural consistency of $\thresmialgo$]\label{thm:corr_mi_cond}
For structure learning  of the Gaussian graphical  model on a graph $G_p \sim \Gmsc(p;\eta,\gamma)$,   $\thresmialgo(\bfx^n;\xi_{n,p},\eta)$ is consistent for a.e. graph $G_p$. In other words,  \beq \lim_{\substack{n,p\to \infty\\ n=\Omega(J_{\min}^{-4}\log p) }}P\left[\thresmialgo \left(\{\bx^n\};\xi_{n,p},\eta \right) \neq G_p \right]=0 \eeq  \et

The proof of this theorem is provided in Section~\ref{sec:cmi}.

\vspace{1em}

\noindent{\bf Remarks: }

\ben \item  For Gaussian random variables, conditional covariances and conditional mutual information are equivalent tests for conditional independence. However, from above results,  we note  that there is a difference in the  sample complexity for the two tests. The sample complexity of $\thresmialgo$ is $n =\Omega(J^{-4}_{\min}\log p)$ in contrast to $n =\Omega(J^{-2}_{\min}\log p)$ for $\threscondalgo$. This is due to faster decay of conditional mutual information on the edges compared to the decay of conditional covariances. Thus, conditional covariances are more efficient for Gaussian graphical model selection compared to conditional mutual information.

\een
\section{Necessary Conditions for Model Selection}\label{sec:necessary}

In the previous sections, we proposed and analyzed efficient algorithms for learning the structure of Gaussian graphical models   Markov on graph ensembles satisfying local-separation property. In this section, we study the   problem of deriving {\em necessary} conditions for consistent structure learning.

For the class of degree-bounded graphs $\Gmsc_{\Deg}(p,\Delta)$, necessary conditions on sample complexity have been characterized before~\citep{Wang&etal:10ISIT} by considering a certain (limited)  set of   ensembles. However, a na\"{i}ve application of such bounds (based on Fano's inequality \citep[Ch. 2]{Cover&Thomas:book})   turns out to be too weak for    the class  of    Erd\H{o}s-R\'{e}nyi  graphs  $\Gmsc_{\ER}(p,c/p)$, where the average degree\footnote{The  techniques in this section are applicable when the average degree  ($c$) of  $\Gmsc_{\ER}(p,c/p)$ ensemble  is a function of $p$, e.g., $c=O(\poly( \log p))$.} $c$ is much smaller than the maximum degree.

We now provide   necessary conditions on the sample complexity for recovery of Erd\H{o}s-R\'{e}nyi  graphs. Our  information-theoretic  techniques may also be applicable to other ensembles of random graphs. This is a promising avenue for future work.

\subsection{Setup}
We now describe the problem more formally. A graph $G$ is drawn from the ensemble of  Erd\H{o}s-R\'{e}nyi  graphs  $G\sim\Gmsc_{\ER}(p,c/p)$.  The learner is also provided with $n$ conditionally i.i.d.\ samples $\bX^n:=(\bX_1 ,\ldots, \bX_n )\in (\calX^p)^n$ (where $\calX = \R$) drawn from the  conditional (Gaussian) product probability density function (pdf) $\prod_{i=1}^n f(\bx_i|G)$. The task is then to estimate $G$, a random quantity. The estimate is denoted as $\hG:=\hG(\bX^n)$.  It is desired to derive tight necessary conditions on $n$ (as a function of $c$ and $p$) so that the {\em probability of error}
\begin{equation}
\Pep:=P( \hG \ne G) \to 0  \label{eqn:consistency}
\end{equation}
as the number of nodes $p$ tends to infinity.
Note that the probability measure $P$ in~\eqref{eqn:consistency} is associated to {\em both} the realization of the random graph $G$ and the samples $\bX^n$.
%, i.e., the measure $P$ can be decomposed as the product measure $P_G \times \prod_{i=1}^n P_{\bX_i|G}$.

\begin{figure}
\centering
\begin{picture}(200,40)
\put(0,15){\vector(1,0){30}}
\put(80,15){\vector(1,0){30}}
\put(160,15){\vector(1,0){30}}

\put(30,0){\line(1,0){50}}
\put(30,0){\line(0,1){30}}
\put(80,0){\line(0,1){30}}
\put(30,30){\line(1,0){50}}

\put(110,0){\line(1,0){50}}
\put(110,0){\line(0,1){30}}
\put(160,0){\line(0,1){30}}
\put(110,30){\line(1,0){50}}

\put(-40, 22){\mbox{$G\sim \Gmsc_{\ER}(p,\frac{c}{p})$}}
\put(33, 14){\mbox{\small $\displaystyle \prod_{i=1}^n f   (\bx_i|G) $}}
\put(77, 35){\mbox{$\bX^n \in (\R^p)^n$}}
\put(120, 14){\mbox{Decoder}}
\put(180, 22){\mbox{$\hG$}}
\end{picture}
\caption{The estimation problem is analogous to source coding: the ``source'' is $G\sim  \Gmsc_{\mathrm{ER}}(p,\frac{c}{p})$, the ``message'' is $\bX^n \in (\R^p)^n$ and the ``decoded source'' is $\hG$. We are asking what the minimum ``rate'' (analogous to the number of samples $n$) are required so that $\hG=G$ with high probability.}
\label{fig:source_code}
\end{figure}
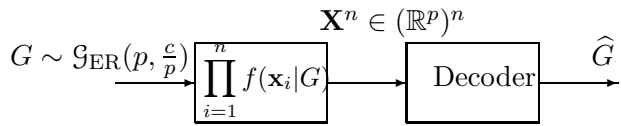

The task is reminiscent of source coding (or compression), a problem of central importance in information theory~\citep{Cover&Thomas:book} -- we would like to derive fundamental limits associated to the problem of reconstructing the source $G$ given a compressed version  of it  $\bX^n$ ($\bX^n$ is also analogous to the ``message''). However, note the important distinction; while in source coding, the source coder can design both the encoder {\em and} the decoder, our problem mandates that the code is fixed by the conditional probability density $f(\bx|G)$. We are only allowed to design the decoder.  See comparisons in Figs.\ \ref{fig:source_coding} and  \ref{fig:source_code}.

\begin{figure}
\centering
\begin{picture}(200,40)
\put(0,15){\vector(1,0){30}}
\put(80,15){\vector(1,0){30}}
\put(160,15){\vector(1,0){30}}

\put(30,0){\line(1,0){50}}
\put(30,0){\line(0,1){30}}
\put(80,0){\line(0,1){30}}
\put(30,30){\line(1,0){50}}

\put(110,0){\line(1,0){50}}
\put(110,0){\line(0,1){30}}
\put(160,0){\line(0,1){30}}
\put(110,30){\line(1,0){50}}

\put(-35, 22){\mbox{$X^m\sim P^m(x)$}}
\put(35, 14){\mbox{Encoder}}
\put(77, 35){\mbox{$M\in [2^{m R}]$}}
\put(120, 14){\mbox{Decoder}}
\put(180, 22){\mbox{$\hX^m$}}
\end{picture}
\caption{The canonical source coding problem. See Chapter 3 in \citep{Cover&Thomas:book}.}
\label{fig:source_coding}
\end{figure}
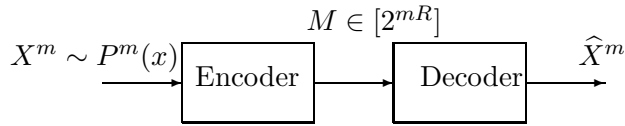

\subsection{Necessary Conditions for Exact Recovery}
To derive the necessary condition for learning Gaussian graphical models Markov on  sparse  Erd\H{o}s-R\'{e}nyi  graphs   $G\sim\Gmsc_{\ER}(p,c/p)$, we assume that the strict walk-summability condition with parameter $\alpha$, according to \eqref{eqn:walksummable}.  We are then able to demonstrate the following:
\begin{theorem}[Weak Converse for Gaussian Models] \label{thm:converse_gaussian}
For a walk-summable Gaussian graphical model satisfying \eqref{eqn:walksummable} with parameter $\alpha$, for almost every graph $G\sim \Gmsc_{\ER}(p, c/p)$ as $p\to\infty$, in order for $\Pep  \to 0$, we require that
\begin{equation}
n\ge \frac{2}{p \log_2  \left[2\pi e \left( \frac{1}{1-\alpha}+1 \right) \right]}\binom{p}{2} \Hb \left(\frac{c}{p}\right) \label{eqn:lower_bound_gauss}
\end{equation}
for all $p$ sufficiently large.
%then, $P( \hG(\bx^m )\ne G)\ge 1-2\epsilon$ for all $n$ sufficiently large for any estimator $\hG(\fndot)$.
\end{theorem}
The proof is provided in Section~\ref{prf:nec}.  By expanding the binary entropy function, it is easy to see that the statement in \eqref{eqn:lower_bound_gauss} can be weakened to the necessary condition:
\begin{equation}
n\ge \frac{  c\log_2 p}{    \log_2  \left[2\pi e \left( \frac{1}{1-\alpha}+1 \right) \right]}. \label{eqn:lower_bound_alt_gauss}
\end{equation}
The above condition   does not involve any asymptotic notation, and also demonstrates   the dependence of the sample complexity on $p,c$ and  $\alpha$   transparently.  Finally, the dependence on $\alpha$ can be explained as follows:    any $\alpha$-walk-summable model is also $\beta$-walk-summable for all $\beta>\alpha$. Thus, the class of $\beta$-walk-summable models   contains the class of  $\alpha$-walk-summable models. This results in a looser bound in \eqref{eqn:lower_bound_gauss} for larger $\alpha$.

%Note that as $\alpha\uparrow 1$, the required number of samples in \eqref{eqn:lower_bound_alt_gauss} decreases.  This is because the set of admissible $\bX^n$ is large (resp.\ small) if $\alpha\approx 1$ (resp.\ if $\alpha\approx 0$). As a result, the effective coding rate decreases as $\alpha\uparrow 1$.

 \subsection{Necessary Conditions for Recovery with Distortion}
In this section,  we generalize Theorem \ref{thm:converse_gaussian} to the case where we only require estimation of the underlying graph up to a certain edit distance: an error is declared if and only if the estimated graph $\hG$ exceeds an edit distance (or distortion) $D$ of the true graph. The {\em edit distance}  $d:\frakG_p\times \frakG_p\to\Nbb\cup\{0\}$  between two undirected graphs $G =(V,E)$ and $G=(V, E')$ is defined as $d(G,G') := |E\triangle E'|$, where $\triangle$ denotes the symmetric difference between the edge sets $E$ and $E'$. The edit distance can be regarded as a distortion measure between two graphs.

Given an positive integer $D$, known as the {\em distortion}, suppose we declare an error if and only if 	 $d(G,G')>D$, then  the probability of error is redefined as
\begin{equation}
\Pep:=P( d(G, \hG(\bX^n)) > D).  \label{eqn:perr_new}
\end{equation}
We derive necessary conditions on $n$ (as a function of $p$ and $c$)  such that  the probability of error~\eqref{eqn:perr_new}  goes to zero as $p\to\infty$. To ease notation, we define the ratio
\begin{equation}
\beta := { D}/ {\binom{p}{2}}. \label{eqn:alpha_def}
\end{equation}
Note that $\beta$ may be  a function of $p$. We do not attempt to make this dependence explicit. The following corollary is based on an idea propounded by~\citet{Kim08} among others.

\begin{corollary}[Weak Converse for Discrete Models With Distortion] \label{cor:converse_gauss_gen}
For $\Pep \to 0$, we must have
\begin{equation}
n\ge \frac{2}{p   \log_2  \left[2\pi e \left( \frac{1}{1-\alpha}+1 \right) \right] } \binom{p}{2} \left[ \Hb \left(\frac{c}{p}\right) -\Hb \left( \beta \right) \right] \label{eqn:lower_bound_general}
\end{equation}
for all $p$ sufficiently large.
%then, $P( \hG(\bx^m )\ne G)\ge 1-2\epsilon$ for all $n$ sufficiently large for any estimator $\hG(\fndot)$.
\end{corollary}
The proof of this corollary is provided in Section~\ref{sec:proof_distort}.  Note that for  \eqref{eqn:lower_bound_general} to be a useful bound, we need $\beta<c/p$ which translates to an allowed distortion $D < cp/2$. We observe from  \eqref{eqn:lower_bound_general} that because the error criterion has been relaxed, the required number of samples is also reduced from the corresponding lower bound in \eqref{eqn:lower_bound_gauss}.

\subsection{Proof Techniques}
Our analysis tools for deriving necessary conditions for Gaussian graphical model selection are information-theoretic in nature. A common and natural tool to derive necessary conditions (also called converses) is to resort to Fano's inequality \citep[Chapter 2]{Cover&Thomas:book}, which (lower) bounds the probability of error $\Pep$ as a function of the {\em equivocation} or {\em conditional entropy} $H(G|\bX^n)$ and the size of the set of all graphs with $p$ nodes. However, a direct and na\"{i}ve application Fano's inequality results in a trivial lower bound as the set of all graphs, which can be realized by $\Gmsc_{\ER}(p,c/p)$ is, loosely speaking, ``too large''.

To ameliorate such a problem, we employ another information-theoretic notion, known as {\em typicality}. A {\em typical set} is, roughly speaking, a set that has small cardinality and yet has  high probability as $p \to \infty$. For example, the probability of a set of length-$m$ sequences is of the order $\approx 2^{mH}$ (where $H$ is the entropy rate of the source) and hence those sequences with probability close to this value are called {\em typical}. In our context, given a graph $G$, we define the   $\bar{d}(G)$ to be the ratio of the number of edges of $G$ to the total number of nodes $p$.  Let $\frakG_p$ denote the set of all graphs with $p$ nodes. For a fixed $\epsilon>0$, we define the following set of graphs:\begin{equation}
\calT_{\epsilon}^{(p)} := \left\{G\in\frakG_p: \left| \frac{\bar{d}(G)}{c} - \frac{1}{2} \right| \le \frac{\epsilon}{2}  \right\}.
\end{equation}
The set $\calT_{\epsilon}^{(p)}$ is known as the {\em $\epsilon$-typical set  of graphs}.  Every graph $G\in \calT_{\epsilon}^{(p)}$ has an average number of edges that is $\frac{c}{2}\epsilon$-close   in the Erd\H{o}s-R\'{e}nyi ensemble. Note that typicality ideas are usually used to derive {\em sufficient} conditions in information theory~\citep{Cover&Thomas:book} ({\em achievability} in information-theoretic parlance); our use of {\em both} typicality  for graphical model selection as well as Fano's inequality to derive converse statements seems novel. Indeed, the proof of the converse of the source coding theorem in \citet[Chapter 3]{Cover&Thomas:book} utilizes only Fano's inequality. We now summarize the properties of the typical set.

\begin{lemma}[Properties of $\calT_{\epsilon}^{(p)}$]\label{lem:typ}
The $\epsilon$-typical set of graphs has the following properties:
\begin{enumerate}
\item $P(\calT_{\epsilon}^{(p)}) \to 1$ as $p\to\infty$.
\item For all $G\in\calT_{\epsilon}^{(p)}$, we have\footnote{We use the notation $\exp_2(\fndot)$ to mean $2^{(\fndot)}$.}
\begin{equation}
\exp_2\left[- \binom{p}{2} \Hb \left(\frac{c}{p}\right)   (1+\epsilon)  \right]\le  P(G)\le \exp_2\left[ - \binom{p}{2} \Hb \left(\frac{c}{p}\right) \right]. \label{eqn:prob}
\end{equation}
\item The cardinality of the  $\epsilon$-typical set can be bounded as
\begin{equation}
(1-\epsilon)\exp_2\left[   \binom{p}{2} \Hb \left(\frac{c}{p}\right) \right] \le |\calT_{\epsilon}^{(p)}| \le  \exp_2\left[   \binom{p}{2} \Hb \left(\frac{c}{p}\right)(1+\epsilon) \right] \label{eqn:card}
\end{equation}
for all $p$ sufficiently large.
\end{enumerate}
\end{lemma}
The proof of this lemma can be found in Section~\ref{prf:lem:typ}.  Parts 1 and 3 of Lemma~\ref{lem:typ} respectively  say  that the set of typical graphs has high probability   and    has very small cardinality relative to the number of graphs with $p$ nodes  $|\frakG_p|=\exp_2(\binom{p}{2})$.    Part 2 of Lemma~\ref{lem:typ} is known as the {\em asymptotic equipartition property}: the graphs in the typical set are almost uniformly distributed.

\section{Implications on Loopy Belief Propagation}\label{sec:lbp}

An active area of research in the graphical model community is that of inference -- i.e.,   the task of computing   node marginals (or MAP estimates) through efficient distributed algorithms. The simplest of these algorithms is the belief propagation\footnote{The variant of the belief propagation algorithm which computes the MAP estimates is known as the max-product
algorithm.} (BP) algorithm, where messages are passed
among the neighbors of the  graph of the model. It is known that belief propagation (and max-product) is exact on tree models, meaning that correct marginals  are computed at all the nodes~\citep{Pearl:book}. On the other hand on general graphs, the generalized version of BP, known as loopy belief propagation (LBP), may not converge and even if it does, the marginals may not be correct.  Motivated by the twin problems of convergence and correctness, there has been   extensive work on characterizing LBP's  performance for different models. See Section~\ref{sec:related_lbp} for details.
As a by-product of our previous analysis on graphical model selection, we now show the asymptotic correctness of  LBP on walk-summable Gaussian models when the underlying graph is   locally tree-like.

\subsection{Background}

The belief propagation (BP) algorithm is a  distributed algorithm where messages  (or beliefs)  are passed
among the neighbors to draw inferences at the nodes of a graphical model. The computation of node marginals through na\"{i}ve variable elimination (or Gaussian elimination in the Gaussian setting) is prohibitively  expensive. However, if the graph is sparse (consists of few edges), the computation of  node marginals may be sped up dramatically  by exploiting the graph structure and using distributed algorithms to parallelize the computations.

For the sake of completeness, we now recall the basic steps in  LBP, specific to Gaussian graphical models.
Given a message schedule which specifies how messages are exchanged, each node $j$  receives information from each of its
neighbors (according to the graph), where the message, $m^t_{i\rightarrow j}(x_j)$, from   $i$ to $j$, in $t^{\tha}$ iteration is parameterized as\[m^t_{i\rightarrow j}(x_j):=\exp\left[-\frac{1}{2}\Delta J^t_{i \rightarrow j}x_j^2+\Delta h^t_{i \rightarrow j} x_j\right].\]Each node $i$ prepares message $m^t_{i\rightarrow j}(x_j)$ by collecting messages from neighbors of the previous iteration (under parallel iterations), and computing \[ \hat{J}_{i\backslash j}(t)= J(i,i)+ \sum_{k\in \nbd(i)\setminus j}\Delta J^{t-1}_{k\rightarrow i}, \quad \hat{h}_{i\backslash j}(t)=h(i)+\sum_{k\in \nbd(i)\setminus j}\Delta h_{k\rightarrow i}(t),\] where \[\Delta J_{i\rightarrow j}^t = - J(j,i)\hat{J}_{i\backslash j}^{-1}(t) J(j,i),\quad \Delta h^t_{i\rightarrow j}=- J(j,i)\hat{J}_{i\backslash j}^{-1}(t)\hat{h}_{k\rightarrow i}(t).\]

\subsection{Results}

Let $\Sigma_{\lbp}(i,i)$ denote the variance at node $i$ at the LBP fixed point.\footnote{Convergence of LBP on walk-summable models has been established by~\citet{Malioutov&etal:06JMRL}.} Without loss of generality, we consider the normalized version of the precision matrix \[ \bfJ=\bfI - \bfR,\] which can always be obtained from a general precision matrix via normalization. We can then renormalize the variances, computed  via LBP, to obtain the variances corresponding to the unnormalized precision matrix.

We consider the following ensemble of locally-tree like graphs. Consider the event that the neighborhood of a node $i$ has no cycles up to graph distance $\gamma$, given by \[\Gamma(i;\gamma,G):=\{B_\gamma(i; G)\mbox{ does not contain any cycles}\}.\] We assume  a random graph  ensemble $\Gmsc(p)$ such that for a given  node $i\in V$,   we have \beq \label{eqn:cyclethreshold}P[\Gamma^c(i;\gamma,G)] = o(1).\eeq

\begin{proposition}[Correctness of LBP]\label{corr:lbp}Given an $\alpha$-walk-summable Gaussian graphical model on  a.e. locally tree-like graph $G\sim \Gmsc(p;\gamma)$ with parameter $\gamma$ satisfying \eqref{eqn:cyclethreshold},  we have  \beq  |\Sigma_G(i,i)-\Sigma_{\lbp}(i,i)|\overset{\mbox{a.a.s.}}{=} O(\max(\alpha^\gamma,P[\Gamma^c(i;\gamma,G)]) ). \eeq\end{proposition}

The proof is given in Section~\ref{proof:lbp}.

\vspace{1em}

\noindent{\bf Remarks: }

\ben \item The class of  Erd\H{o}s-R\'{e}nyi random graphs, $G\sim \Gmsc_{\ER}(p,c/p)$ satisfies \eqref{eqn:cyclethreshold}, with $\gamma = O(\log p/\log c)$ for a node $i \in V$ chosen uniformly at random.

\item Recall that the class of random regular graphs $G\sim \Gmsc_{\reg}(p,\Delta)$ have a girth of $O(\log_{\Delta-1} p)$. Thus, for any node $i \in V$, \eqref{eqn:cyclethreshold} holds with $\gamma=O(\log_{\Delta-1} p)$.
 \een

\subsection{Previous Work on Loopy Belief Propagation}\label{sec:related_lbp}

It has long been known through numerous empirical studies~\citep{murphy1999loopy}
and the phenomenal successes of turbo decoding~\citep{mceliece2002turbo}, that loopy belief propagation
(LBP) performs reasonably well on a variety of graphical models though it also must be
mentioned that LBP fails catastrophically on other models. ~\citet{Weiss:00Neural}
proved that if the underlying graph (of a Gaussian graphical model) consists of a
single cycle, LBP converges and is correct, i.e., the fixed points of the means and
the variances are the same as the true means and variances. In addition, sufficient
conditions for a unique fixed point are known~\citep{mooij2007sufficient}. The max-product variant of LBP
(called the max-product or min-sum algorithm) has been studied~\citep{Bayati&etal:05ISIT,sanghavi2009message,Ruozzi&Tatikonda:10Arxiv}. Despite its
seemingly heuristic nature, LBP has found a variety of concrete applications, especially
in combinatorial optimization~\citep{moallemi2010convergence,gamarnik2010belief}. Indeed, it has been applied and analyzed for
NP-hard problems such as maximum matching~\citep{bayati2008max}, b-matching~\citep{sanghavi2009message}, the Steiner tree
problem~\citep{bayati2008rigorous}.

The application of BP for inference in Gaussian graphical models has been studied
extensively -- starting with the seminal work by~\citet{Weiss&Freeman:01Neural}. Undoubtedly
the Kalman filter is the most familiar instance of BP in Gaussian graphical models. The
notion of walk-summability in Gaussian graphical models was introduced by~\citet{Malioutov&etal:06JMRL}. Among
other results, the authors showed that LBP converges to the correct means for walk-
summable models but the estimated variances may nevertheless still be incorrect. \citet{Chandrasekaran:08SP} leveraged the ideas of~\citet{Malioutov&etal:06JMRL} to analyze related inference algorithms such
as embedded trees and the block Gauss-Seidel method. Recently,~\citet{Liu&etal:10ISIT}
considered a modified version of LBP by identifying a special set of nodes -- called the
feedback vertex set (FVS) \citep{Vazirani:book} -- that breaks (or approximately breaks)
cycles in the loopy graph. This allows one to perform inference in a tractably to tradeoff
accuracy and computational complexity. For Gaussian graphical models Markov on
locally tree-like graphs, an {\em approximate} FVS can be identified. This set, though
not an FVS {\em per se}, allows one to break all the short cycles in the graph and thus, it
allows for proving tight error bounds on the inferred variances. The performance of LBP
on locally tree-like graphs has also been studied for other families of graphical models.
For Ising models Markov on locally tree-like graphs, the work by~\citet{Dembo&Montanari:08Arxiv} established an
analogous result for attractive (also known as ferromagnetic) models. Note that walk-summable Gaussian graphical models is a superset of the class of attractive Gaussian
models. An interpretation of LBP in terms of graph covers is given by~\citet{Vontonbel:10IT} and its equivalence to walk-summability for Gaussian graphical models is established by~\citet{ruozzi2009graph}.

\section{Conclusion}\label{sec:conclusion}

In this paper, we adopted a novel and a unified paradigm for graphical model selection. We presented a simple local algorithm for structure estimation with low computational and sample complexities under a set of mild and transparent conditions. This algorithm succeeds on a wide range of graph ensembles such as the Erd\H{o}s-R\'{e}nyi ensemble, small-world networks etc.  We also employed novel information-theoretic techniques for establishing necessary conditions for graphical model selection.

\subsubsection*{Acknowledgement}

The first author is supported in part by the setup funds at UCI and the AFOSR Award FA9550-10-1-0310, the second author is supported by A*STAR, Singapore and the third author is supported in part by AFOSR under Grant FA9550-08-1-1080.
The authors thank Venkat Chandrasekaran (MIT) for discussions on walk-summable models, Elchanan Mossel (UC Berkeley) for discussions on the necessary conditions for model selection and Divyanshu Vats (U. Minn.) for extensive comments. The authors thank the Associate Editor Martin Wainwright (Berkeley) and the anonymous reviewers for comments which significantly improved this manuscript.

\begin{appendix}

\section{Walk-summable Gaussian  Graphical Models}\label{sec:walkdiscuss}

\subsection{Background on Walk-Summability}\label{sec:backgroundwalk}

We now recap the properties   of  walk-summable Gaussian graphical models, as given by \eqref{eqn:walksummable}. For details, see~\citet{Malioutov&etal:06JMRL}.
For simplicity, we first  assume that the  diagonal of the  potential matrix $\bfJ$ is normalized   ($J(i,i)=1$ for all $i \in V$). We remove this assumption and consider general unnormalized precision matrices in Section~\ref{sec:unnormalized}. Consider  splitting the matrix $\bfJ$ into the identity matrix  and the partial correlation  matrix $\bfR$, defined in \eqref{eqn:parcor}: \beq\label{eqn:potential_normalized}\bfJ= \bfI-\bfR.\eeq
The covariance matrix $\Sigmabf$ of the graphical model in \eqref{eqn:potential_normalized} can be decomposed as \beq\label{eqn:covseries} \Sigmabf= \bfJ^{-1}= (\bfI-\bfR)^{-1} = \sum_{k=0}^\infty \bfR^k, \quad \norm{\bfR}<1,\eeq using Neumann power series for the matrix inverse. Note that we require  that $\norm{\bfR}<1$ for \eqref{eqn:covseries} to hold, which is implied by walk-summability in \eqref{eqn:walksummable} (since $\norm{\bfR}\leq \norm{\overline{\bfR}}$).

We now relate the matrix power $\bfR^l$ to walks on graph $G$. A walk $\bfw$ of length $l\geq 0$ on graph $G$ is a sequence of nodes  $\bfw:=(w_0, w_1, \ldots, w_l)$ traversed on the graph $G$, i.e., $(w_k, w_{k+1})\in G$. Let $|\bfw|$ denote the length of the walk. Given matrix $\bfR_G$ supported on graph $G$, let the weight of the walk be\[ \phi(\bfw):= \prod_{k=1}^{|\bfw|}R(w_{k-1}, w_k).\] The elements of the matrix power $\bfR^l$ are given by \beq \label{eqn:matrix_power_walk} R^l(i,j) = \sum_{\bfw:i \overset{l}{\rightarrow} j} \phi(\bfw),\eeq where $i \overset{l}{\rightarrow} j$ denotes the set of  walks from $i$ to $j$ of length $l$. For this reason, we henceforth refer to $\bfR$ as the {\em walk matrix}.

Let $i \rightarrow j$ denote all the walks between $i$ and $j$.
Under the walk-summability condition in \eqref{eqn:walksummable}, we have convergence of $   \sum_{\bfw:i \rightarrow j}  \phi(\bfw) $, irrespective of the order in which the walks are collected, and this  is equal to the covariance $\Sigma(i,j)$.

In Section~\ref{sec:covbound}, we   relate walk-summability in \eqref{eqn:walksummable} to the notion of correlation decay, where the effect of faraway nodes on covariances can be controlled and the local-separation property of the graphs under consideration can be exploited.

%In turn we can consider summing all walks between $i$ and $j$ \[  \sum_{\bfw:i \rightarrow j} \phi(\bfw),\] which converges when the walks are ordered by increasing length

\subsection{Sufficient Conditions for Walk-summability}\label{sec:walkrandom}

We now provide sufficient conditions and suitable parameterization for walk-summability in \eqref{eqn:walksummable} to hold.  The adjacency matrix $\bfA_G$ of a graph $G$ with maximum degree $\Delta_G$ satisfies\[ \lambda_{\max}(\bfA_G) \leq \Delta_G,\] since it is dominated by a $\Delta$-regular graph which has maximum eigenvalue of $\Delta_G$. From Perron-Frobenius theorem, for adjacency matrix $\bfA_G$, we have $\lambda_{\max}(\bfA_G) = \norm{\bfA_G}$, where $\norm{\bfA_G}$ is the spectral radius of $\bfA_G$.
Thus, for $\overline{\bfR}_G$ supported on graph $G$, we have
\[\alpha:=\norm{\overline{\bfR}_G} =O\left(J_{\max} \Delta\right)
,\] where $J_{\max}:= \max_{i,j} |R(i,j)|$. This implies that \beq J_{\max} =O\left(\frac{1}{\Delta}\right)\eeq to have $\alpha<1$, which  is the requirement for walk-summability.

When the graph  $G$ is a Erd\H{o}s-R\'{e}nyi random graph, $G\sim \Gmsc_{\ER}(p,c/p)$, we can provide better bounds.
When $G\sim \Gmsc_{\ER}(p,c/p)$, we have~\citep{Krivelevich:03Prob}, that
\[ \lambda_{\max}(\bfA_G) = (1+o(1))\max(\sqrt{\Delta_G}, c),\] where $\Delta_G$ is the maximum degree and $\bfA_G$ is the adjacency matrix. Thus, in this case, when $c=O(1)$, we require that \beq J_{\max} =O\left(\sqrt{\frac{1}{\Delta}}\right),\eeq for walk-summability $(\alpha<1)$.
Note that when $c=O(\poly(\log p))$, w.h.p. $\Delta_{G_p} = \Theta(\log p / \log \log p)$~\citep[Ex. 3.6]{Bollobas:book}.

%Similarly, for power-law graphs with average degree $c$,
%A sub-class of walk-summable matrices are the diagonally dominant matrices. We can parameterize them as  \bcase{R(i,j)=}\frac{\delta_{i,j}}{\sqrt{\Deg(i) \Deg(j)}}, & $(i,j)\in G$\label{eqn:Rnorm}\\ 0& o.w. \ecase In this case, we have $\norm{\overline{\bfR}_G}\leq \alpha:= \max_{i,j}|\delta_{i,j}| $ from Rayleigh quotient theorem~\cite{Chung:book2}, for any graph\footnote{The parameterization in \eqref{eqn:Rnorm} is on the lines of  the normalized laplacian matrix $\bfL_G$ for graph $G$~\cite{Chung:book2}.} $G$. This in turn implies that $\lambda_{\min}(\bfJ_G)\geq 1- \delta_{\max} >0$ for $0<\delta_{\max}<1$.
%Thus, by suitably normalizing for the node degrees, we can obtain diagonally dominant (and thus walk-summable) matrices.

\subsection{Implications of Walk-Summability}\label{sec:covbound}

Recall that $\Sigmabf_G$ denotes the covariance matrix for Gaussian graphical model on graph $G$ and that $\bfJ_G=\Sigmabf^{-1}_G$ with $\bfJ_G= \bfI-\bfR_G$ in \eqref{eqn:potential_normalized}. We now relate the walk-summability condition in \eqref{eqn:walksummable}
to correlation decay in the model. In other words, under walk-summability, we can show that the effect of faraway nodes on covariances decays with distance, as made precise in Lemma~\ref{lemma:covbound}.

Let $B_\gamma(i)$ denote the set of nodes within $\gamma$ hops from node $i$ in graph $G$.
Denote \beq \label{eqn:H}H_{\gamma;ij}:=G(B_\gamma(i)\cap B_\gamma(j))\eeq as the induced subgraph of $G$ over the intersection of $\gamma$-hop neighborhoods at $i$ and $j$ and retaining the nodes  in $V\setminus \{B_{\gamma}(i)\cap B_{\gamma}(j)\}$. Thus, $H_{\gamma; ij}$ has the same number of nodes as $G$. .
We first make the following simple observation: the  $(i,j)$ element in the $\gamma^{\tha}$ power of walk  matrix,  $R_G^\gamma(i,j)$, is given by walks of length   $\gamma$ between $i$ and $j$ on graph $G$ and thus, depends only on subgraph\footnote{Note that $R^\gamma(i,j)=0$ if $B_\gamma(i)\cap B_\gamma(j)=\emptyset$.}  $H_{\gamma;ij}$ (see \eqref{eqn:matrix_power_walk}). This   enables us to quantify the effect of nodes outside  $B_\gamma(i)\cap B_\gamma(j)$ on the covariance $\Sigma_G(i,j)$.

Define a new walk matrix $\bfR_{H_{\gamma;ij}}$   such that \bcase{R_{H_{\gamma;ij}}(a,b) = }R_{G}(a,b),& $a,b\in  B_\gamma(i)\cap B_\gamma(j)$,\\  0,  &o.w.\ecase In other words, $\bfR_{H_{\gamma;ij}}$ is formed by considering the Gaussian graphical model over graph $H_{\gamma;ij}$.
Let $\Sigmabf_{H_{\gamma;ij}}$ denote the corresponding covariance matrix.\footnote{When $B_\gamma(i)\cap B_\gamma(j)=\emptyset$ meaning that graph distance between $i$ and $j$ is more than $\gamma$, we obtain $\Sigmabf_{H_{\gamma;ij}}=\bfI$.}

\begin{lemma}[Covariance Bounds Under Walk-summability]\label{lemma:covbound}For any walk-summable Gaussian graphical model    $(\alpha:=\norm{\overline{\bfR}_G}<1)$, we have\footnote{The bound in \eqref{eqn:covbound} also holds if $H_{\gamma;ij}$ is replaced with any of its supergraphs.}\beq\label{eqn:covbound} \max_{i,j} |\Sigma_G(i,j)-\Sigma_{H_{\gamma;ij}}(i,j)|\leq \alpha^\gamma\frac{2 \alpha}{1-\alpha}= O(\alpha^\gamma).\eeq
\end{lemma}

Thus, for  walk-summable Gaussian graphical models, we have $\alpha:=\norm{\overline{\bfR}_G}<1$, implying that the error in \eqref{eqn:covbound} in approximating the covariance by local neighborhood decays exponentially with distance.
Parts of the proof below  are inspired by~\citet{Dumitriu:09Arxiv}.

\bprf
Using the power-series in \eqref{eqn:covseries}, we can write the covariance matrix as
\[ \Sigmabf_G = \sum_{k=0}^\gamma \bfR_G^k + \bfE_G,\] where the error matrix $\bfE_G$ has spectral radius \[\norm{\bfE_G} \leq \frac{\norm{\bfR_G}^{\gamma+1}}{1-\norm{\bfR_G}}, \] from \eqref{eqn:covseries}. Thus,\footnote{For any matrix $\bfA$, we have $\max_{i,j} |A(i,j)| \leq \norm{\bfA}$.} for  any  $i,j \in V$,
\beq\label{eqn:error1} |\Sigma_G(i,j) - \sum_{k=0}^\gamma R^k_G(i,j)|\leq \frac{\norm{\bfR_G}^{\gamma+1}}{1-\norm{\bfR_G}}.\eeq
Similarly, we have   \begin{align} |\Sigma_{H_{\gamma;ij}}(i,j)-\sum_{k=0}^\gamma R_{H_{\gamma;ij}}^k(i,j)| &\leq \frac{\norm{\bfR_{H_{\gamma;ij}}}^{\gamma+1}}{1-\norm{\bfR_{H_{\gamma;ij}}}}\\ &\lea \frac{\norm{\overline{\bfR}_G}^{\gamma+1}}{1-\norm{\overline{\bfR}_G}} ,\label{eqn:error2}\end{align} where for inequality (a), we use the fact that \[\norm{\bfR_{H_{\gamma;ij}}} \leq \norm{\overline{\bfR}_{H_{\gamma;ij}}}\leq \norm{\overline{\bfR}_G},\]since $H_{\gamma;ij}$ is a subgraph\footnote{When two matrices $\bfA$ and $\bfB$ are such that  $\abs{A(i,j)}\geq \abs{B(i,j)}$ for all $i,j$, we have $\norm{\bfA}\geq \norm{\bfB}$.} of $G$.

Combining \eqref{eqn:error1}  and \eqref{eqn:error2},  using the triangle inequality,  we obtain \eqref{eqn:covbound}.\eprf

We also make some simple observations about conditional covariances in walk-summable models. Recall that  $\overline{\bfR}_G$ denotes matrix with absolute values of $\bfR_G$, and $\bfR_G$ is the walk matrix over graph $G$. Also recall that the $\alpha$-walk summability condition in \eqref{eqn:walksummable}, is $  \norm{\overline{\bfR}_{G}}\leq \alpha<1$.

\begin{proposition}[Conditional Covariances under Walk-Summability]\label{prop:condcov}Given a  walk-summable Gaussian graphical model, for any $i,j\in V$ and $S\subset V$ with $i,j\notin S$, we have \beq\Sigma(i,j|S) = \sum_{\substack{\bfw:i\rightarrow j\\ \forall k \in \bfw, k\notin S}}\phi_G(\bfw).\label{eqn:condcov}\eeq  Moreover, we have \beq \sup_{\substack{i\in V\\ S\subset V\setminus i}}\Sigma(i,i|S) \leq (1-\alpha)^{-1}=O(1) .\label{eqn:linfvar}\eeq\end{proposition}

\bprf We have, from \citet[Thm. 2.5]{Rue&Held:book}, \[ \Sigma(i,j|S) =  J_{-S,-S;G}^{-1}(i,j) ,\] where $\bfJ_{-S,-S;G}$ denotes the submatrix of potential matrix $\bfJ_G$ by deleting nodes in $S$. Since submatrix of a walk-summable matrix is walk-summable, we have \eqref{eqn:condcov}  by appealing to the walk-sum expression for conditional covariances.

For \eqref{eqn:linfvar},  let $\infnorm{\bfA}$ denote the maximum absolute value of entries in matrix $\bfA$. Using monotonicity of spectral norm and the fact that $\infnorm{\bfA}\leq \norm{\bfA}$, we have
\begin{align}\sup_{\substack{i\in V\\ S\subset V, i\notin V}}\Sigma(i,i|S) &\leq \norm{\bfJ_{-S,-S;G}^{-1}}=(1-\norm{\bfR_{-S,-S;G}})^{-1}\nn\\ &\leq(1-\norm{\overline{\bfR}_{-S,-S;G}})^{-1}\leq (1-\norm{\overline{\bfR}_{G}})^{-1}=O(1)\nn. \end{align}
\eprf

Thus, the conditional covariance in \eqref{eqn:condcov} consists of walks in the original graph $G$, not passing through nodes in $S$.

\section{Graphs with Local-Separation Property}\label{proof:graphs}

\subsection{Conditional Covariance   between Non-Neighbors: Normalized Case}

We now provide bounds on the conditional covariance for Gaussian graphical models Markov on a graph $G\sim\Gmsc(p;\eta,\gamma)$ satisfying the local-separation property $(\eta,\gamma)$, as per Definition~\ref{def:localpath}.

\begin{lemma}[Conditional Covariance Between Non-neighbors]\label{lemma:cmisep}For a walk-summable Gaussian graphical model, the conditional covariance between non-neighbors $i$ and $j$, conditioned on  $S_\gamma$, the $\gamma$-local separator between $i$ and $j$, satisfies
\beq\max_{j\notin \nbd(i)} \Sigma(i;j| {S_\gamma}) =O(\norm{\overline{\bfR}_G}^{\gamma}).\label{eqn:cmisep}\eeq

%\beq\max_{j\notin \nbd(i)} I(X_i;X_j|\bfX_{S_\gamma}) =O(\rho^{2\gamma}(\overline{\bfR}_G)).\label{eqn:cmisep}\eeq
\end{lemma}

\bprf In this proof, we abbreviate $S_\gamma$ by $S$ for notational convenience.
The conditional covariance is given by the Schur complement, i.e., for any subset $A$ such that  $A\cap S=\emptyset$,\beq\label{eqn:condcovschur} \Sigma(A|S) = \Sigma(A,A)
- \Sigma(A,S) \Sigma(S, S)^{-1} \Sigma(S,A).\eeq

%and the conditional information matrix is given by the submatrix $\bfJ_{A, A;G}$, \[\Cov^{-1}(\bfX_A|\bfX_S)= \bfJ_{A,A;G}.\]

We use the notation $\Sigmabf_G(A, A)$ to denote the submatrix of the covariance matrix $\Sigmabf_G$, when the underlying graph is $G$.
As in Lemma~\ref{lemma:covbound}, we may decompose $\Sigmabf_G$ as follows: \[ \Sigmabf_G  = \Sigmabf_{H_{\gamma}}  + \bfE_\gamma,\] where     $H_\gamma$ is the subgraph spanned by $\gamma$-hop neighborhood $B_\gamma(i)$,   and $\bfE_\gamma$ is the error matrix.
Let $\bfF_\gamma$ be the matrix such that
\[\Sigmabf_G(S,S)^{-1} = \Sigmabf_{H_\gamma}(S,S)^{-1} + \bfF_\gamma.\]

We have $\Sigma_{H_\gamma}(i,j|S) = 0, $ where $\Sigma_{H_\gamma}(i,j|S)$ denotes the conditional covariance  by considering the model given by the subgraph $H_\gamma$. This is due to the Markov property since  $i$ and $j$ are separated by $S$ in the subgraph $H_\gamma$.

Thus using   \eqref{eqn:condcovschur}, the conditional covariance  on graph $G$ can be bounded as \[ \Sigma_G(i,j|S)=O(\max(\norm{\bfE_\gamma}, \norm{\bfF_\gamma})).\] By Lemma~\ref{lemma:covbound}, we have $\norm{\bfE_\gamma}=O(\norm{\overline{\bfR}_G}^\gamma)$. Using   Woodbury   matrix-inversion identity, we also have $\norm{\bfF_\gamma}=O(\norm{\overline{\bfR}_G}^\gamma)$.
\eprf

\subsection{Extension to General Precision Matrices: Unnormalized Case}\label{sec:unnormalized}

We now extend the above analysis to general precision matrices $\bfJ$ where the diagonal elements are not assumed to be identity. Denote the precision matrix as
\[ \bfJ = \bfD - \bfE,\] where $\bfD$ is a diagonal matrix and $\bfE$ has zero diagonal elements. We thus have that
\beq\label{eqn:Jnorm} \bfJ_{\Norm}:= \bfD^{-0.5}\bfJ\bfD^{-0.5} = \bfI - \bfR,\eeq where $\bfR$ is the partial correlation matrix. This also implies that \[ \bfJ = \bfD^{0.5}\bfJ_{\Norm} \bfD^{0.5}.\]Thus, we have that
\beq\label{eqn:Sigmanorm} \Sigmabf = \bfD^{-0.5} \Sigmabf_{\Norm}\bfD^{-0.5}  , \eeq where $\Sigmabf_{\Norm}:= \bfJ_{\Norm}^{-1}$  is the covariance matrix corresponding to the normalized model.  When the model is walk-summable, i.e., $\norm{\overline{\bfR}}\leq \alpha<1$, we have that $\Sigmabf_{\Norm}= \sum_{k \geq 0} \bfR^k$.

We now utilize the results derived in the previous sections involving the normalized model (Lemma~\ref{lemma:covbound} and Lemma~\ref{lemma:cmisep})  to obtain bounds for general precision matrices.

\begin{lemma}[Covariance Bounds for General Models]For any walk-summable Gaussian graphical model    $(\alpha:=\norm{\overline{\bfR}_G}<1)$, we have  the following results:

\ben \item {\bf Covariance Bounds: }The covariance entries upon limiting to a subgraph $H_{\gamma;ij}$ for any $i, j\in V$ satisfies\beq\label{eqn:covboundgen} \max_{i,j} |\Sigma_G(i,j)-\Sigma_{H_{\gamma;ij}}(i,j)|\leq \frac{\alpha^\gamma}{D_{\min} }\frac{2 \alpha}{1-\alpha}= O\left(\frac{\alpha^\gamma}{D_{\min}}\right),\eeq where $D_{\min}:= \min_i D(i,i)=\min_i J(i,i)$.

\item {\bf Conditional Covariance between Non-neighbors: }The conditional covariance between non-neighbors $i$ and $j$, conditioned on  $S_\gamma$, the $\gamma$-local separator between $i$ and $j$, satisfies
\beq\max_{j\notin \nbd(i)} \Sigma(i;j| {S_\gamma})= O\left(\frac{\alpha^\gamma}{D_{\min}}\right),\label{eqn:cmisepgen}
\eeq where $D_{\min}:= \min_i D(i,i)=\min_i J(i,i)$.
 \een\end{lemma}

\bprf Using \eqref{eqn:Sigmanorm} and Lemma~\ref{lemma:covbound}, we have \eqref{eqn:covboundgen}. Similarly, it can be shown that for any $S\subset V\setminus \{i,j\}$, $i,j\in V$, \[ \Sigma(i,j|S)= \bfD^{-0.5} \Sigma_{\Norm}(i,j|S) \bfD^{-0.5},\] where $\Sigma_{\Norm}(i,j|S)$ is the conditional covariance corresponding to the model with normalized precision matrix. From Lemma~\ref{lemma:cmisep}, we have \eqref{eqn:cmisepgen}.  \eprf

\subsection{Conditional Covariance between Neighbors: General Case}\label{sec:nbd}

We  provide a lower bound on conditional covariance among the neighbors for the graphs under consideration. Recall that $J_{\min}$  denotes the minimum   edge potentials.
Let
\[K(i,j):= \norm{\bfJ(V\setminus \{i,j\}, \{i,j\})}^2 , \] where $\bfJ( V\setminus \{i,j\}, \{i,j\})$ is a sub-matrix  of the potential matrix $\bfJ$.
%Recall   the event $\Ac(\gamma;G)$ in \eqref{eqn:eventoverlap}, for $\gamma\geq0$ and graph $G$, is defined as not having any pairs of overlapping cycles in graph $G$ of length at most $\gamma$.

\begin{lemma}[Conditional Covariance Between Neighbors]\label{lemma:positivity}For an $\alpha$-walk summable   Gaussian graphical model satisfying \beq D_{\min}(1-\alpha) \min_{(i,j)\in G_p} \frac{J(i,j)}{K(i,j)}>1+\delta,\label{eqn:minmax2}\eeq
    for some $\delta>0$ (not depending on $p$), where $D_{\min}:=\min_i J(i,i)$, we have
\beq\label{eqn:covminbd}
\abs{\Sigma_G(i,j|S)}= \Omega(J_{\min}),\eeq   for any  $(i,j)\in G$ such that $j \in \nbd(i)$ and any subset $S\subset V$ with $i,j \notin S$.
\end{lemma}

\bprf First note that for attractive models, \begin{align} \nn \Sigma_G(i,j|S) &\gea \Sigma_{G_1}(i,j|S)\\   &\eqb \frac{-J(i,j)}{J(i,i)J(j,j)-J(i,j)^2}=\Omega(J_{\min}), \end{align}  where $G_1$ is the graph consisting only of edge $(i,j)$. Inequality (a) arises from the fact that in attractive models, the weights of  all the walks are positive, and thus, the weight of  walks on $G_1$ form a lower bound for those  on $G$ (recall that the covariances are given by the sum-weight of walks on the graphs).  Equality (b) is   by direct matrix inversion of the model on $G_1$.

%, assuming that $\bfJ=\bfI-\bfR$ is in the normalized form.

For general models, we need further analysis. Let $A=\{i,j\}$ and $B=V\setminus \{S\cup A\}$, for some $S\subset V\setminus A$. Let $\Sigmabf(A,A)$ denote the covariance matrix on set $A$, and let $\tilJbf(A, A) := \Sigmabf(A,A)^{-1}$ denote the corresponding marginal potential matrix. We have for all $S\subset V\setminus A$ \[ \tilJbf(A,A) = \bfJ(A,A) - \bfJ(A,B) \bfJ(B,B)^{-1} \bfJ(B,A).\]
Recall that $\infnorm{\bfA}$ denotes the maximum absolute value of entries in matrix $\bfA$.  \begin{align}\nn \infnorm{\bfJ(A,B) \bfJ(B,B)^{-1} \bfJ(B,A)}&\lea \norm{\bfJ(A,B) \bfJ(B,B)^{-1} \bfJ(B,A)}\\ \nn &\leb  \norm{\bfJ(A,B)}^2\norm{\bfJ(B,B)^{-1}}\\ &= \frac{\norm{\bfJ(A,B)}^2}{\lambda_{\min}(\bfJ(B,B))},
\\ &\lec \frac{K(i,j)^2}{D_{\min}(1-\alpha) }\end{align}
where inequality (a) arises from the  fact that the $\ell_\infty$ norm  is bounded by the spectral norm, (b) arises from sub-multiplicative property of norms and (c) arises from walk-summability property.   Inequality (b) is from the bound on edge potentials and $\alpha$-walk summability of the model and since $K(i,j)\geq\norm{\bfJ(A,B)}$.     Assuming \eqref{eqn:minmax2},  we have \[\abs{\tilJ(i,j)}> J_{\min} - \frac{\norm{\bfJ(A,B)}^2}{D_{\min}(1-\alpha) } = \Omega(J_{\min}).\] Since
\[ \Sigma_G(i,j|S) = \frac{-\tilJ(i,j)}{\tilJ(i,i)\tilJ(j,j)-\tilJ(i,j)^2},\] we have the result.    \eprf

\subsection{Analysis of Loopy Belief Propagation}\label{proof:lbp}

\bprfof{Proposition~\ref{corr:lbp}} From Lemma~\ref{lemma:covbound}  in Section~\ref{sec:covbound}, for any $\alpha$-walk-summable Gaussian graphical model, we have, for all nodes $i\in V$ conditioned on  the   event $\Gamma(i;\gamma,G)$, \beq |\Sigma_G(i,i)-\Sigma_{\lbp}(i,i)|= O(\norm{\overline{\bfR}_G}^\gamma).\eeq
This is because conditioned on $\Gamma(i;\gamma,G)$, it is shown that the series expansions based on walk-sums corresponding to the variances $\Sigma_{H_{\gamma;ij}}(i,i)$   and $\Sigma_{\lbp}(i,i)$ are identical up to length $\gamma$ walks, and the effect of walks beyond length $\gamma$ can be bounded as above. Moreover, for a sequence of $\alpha$-walk-summable, we have $\Sigma(i,i)\leq M$ for all $i\in V$, for some constant $M$ and similarly $\Sigma_{\lbp}(i,j)\leq M'$ for some constant $M'$ since it is obtained by the set of self-avoiding walks in $G$.
We thus have \[\Ebb\left[|\Sigma_G(i,i)-\Sigma_{\lbp}(i,i)|\right]
 \leq  \left[O(\norm{\overline{\bfR}_G}^\gamma) + P[\Gamma^c(i;\gamma)]\right]=o(1),\] where $\Ebb$ is over the expectation of ensemble $\Gmsc(p)$.  By Markov's inequality\footnote{By Markov's inequality, for a non-negative random variable $X$, we have $P[X>\delta] \leq \Ebb[X]/\delta$. By choosing $\delta=\omega(\Ebb[X])$, we have the result.}, we have the result.
\eprfof

\section{Sample-based Analysis}\label{proof:sample}

\subsection{Concentration of Empirical  Quantities}

For our sample complexity analysis, we recap the concentration result by~\citet[Lemma 1]{Ravikumar&etal:08Arxiv} for sub-Gaussian matrices and specialize it to Gaussian matrices.

\bl[Concentration of Empirical Covariances]\label{lemma:miconc}For any $p$-dimensional Gaussian random vector $\bfX=[X_1,\ldots, X_p]$, the empirical covariance obtained from $n$ samples satisfies\beq P\left[  |\hSigma(i,j)-\Sigma(i, j)|> \epsilon\right]\leq 4 \exp\left[- \frac{n\epsilon^2}{ 3200 M^2 }\right],\label{eqn:cov_conc}\eeq for all $\epsilon\in (0,40M)$ and $M:=\max_{i} \Sigma(i,i)$.\el

This translates to bounds for empirical conditional covariance.

\begin{corollary}[Concentration of Empirical Conditional Covariance]\label{lemma:miconc_cond}For a walk-summable $p$-dimensional Gaussian random vector $\bfX=[X_1,\ldots, X_p]$,  we have
\begin{equation}\label{eqn:miconc_cond}
P\left[ \max_{\substack{i\neq j\\ S\subset V, |S|\leq \eta}} |\hSigma(i,j|S) - \Sigma(i;j| S)|>\epsilon\right] \le   4 p^{\eta+2} \exp\left( -\frac{n   \epsilon^2}{  K  } \right),
\end{equation}where   $K\in (0,\infty)$  is a constant which is bounded when   $\norm{\Sigmabf}_\infty$ is bounded, for all $\epsilon\in (0,40M)$ with $M:=\max_{i} \Sigma(i,i)$, and $n\geq \eta$.\end{corollary}

%for all $\epsilon\in (0,40K)$  and

\bprf For a  given $i,j\in V$ and $S\subset V$ with $\eta\leq n$, using \eqref{eqn:condcovschur}, \begin{align}\nn
P\left[  |\hSigma(i,j|S) - \Sigma(i;j| S)|>\epsilon\right] &\leq \Pbb\left[ \left( |\hSigma(i,j) - \Sigma(i;j)|>\epsilon \right)\right.\\ \nn  &\left. \bigcup_{k \in S} \left( |\hSigma(i,k) - \Sigma(i;k)|>K'\epsilon \right) \right] , \end{align}
where $K'$ is a constant which is bounded when $\norm{\Sigmabf}_\infty$ is bounded.  Using
 Lemma~\ref{lemma:miconc}, we have the result. \eprf
%
%\bprf
%Since the model is walk-summable, we have that $M:=\max_{i, S} \Sigma(i,i|S) = O(1)$.   The result then follows from union bound. \eprf

\subsection{Proof of Theorem~\ref{thm:corr_thres_cond}}\label{proof:corr_thres_cond}

We are now ready to prove Theorem~\ref{thm:corr_thres_cond}.
We   analyze the error events for the conditional covariance  threshold test $\threscondalgo$. For  any $(i,j) \notin G_p$, define the event
\beq  \label{eqn:eventcalF1}
\calF_1(i,j; \{\bx^n\} ,G_p):= \left\{ \abs{\hSigma(i,j|S)}>\xi_{n,p}  \right\},
\eeq
where $\xi_{n,p}$ is the threshold in \eqref{eqn:xi} and $S$ is the $\gamma$-local separator between $i$ and $j$ (since the minimum in \eqref{eqn:optsep} is achieved by the $\gamma$-local separator). Similarly for any edge $(i,j)\in G_p$, define the event that
\beq \label{eqn:eventcalF2}
\calF_2(i,j; \{\bx^n\} ,G_p):= \left\{\exists S\subset V: |S|\leq \eta,  \abs{\hSigma(i,j|S)}<\xi_{n,p} \right\}.
\eeq
The probability of error resulting from $\threscondalgo$ can thus be bounded by the two types of errors,
\begin{align}
\label{eqn:twoerrors}
\Pbb [ \threscondalgo(\{\bx^n\};\xi_{n,p
}) \neq G_p] &\leq  \Pbb \left[\bigcup_{(i,j) \in G_p} \calF_2(i,j; \{\bx^n\},G_p) \right]  \nn\\
&\quad+ \Pbb \left[\bigcup_{(i,j) \notin G_p} \calF_1(i,j;  \{\bx^n
\} ,G_p) \right]
\end{align}
For the first term, applying union bound for both the terms and using the result \eqref{eqn:miconc_cond} of Lemma~\ref{lemma:miconc},
\beq \label{eqn:firsterm1}
\Pbb \left[\bigcup_{(i,j) \in G_p} \calF_2(i,j; \{\bx^n\}  ,G_p) \right] =  O\left(p^{\eta+2} \exp\left[-\frac{n (C_{\min}(p)-\xi_{n,p
})^2}{K^2} \right]\right)
\eeq
 where
\begin{equation}\label{eqn:Imin}
C_{\min}(p):= \inf_{\substack{(i,j)\in G_p\\ S\subset V, i,j\notin S\\|S|\leq \eta}}
\abs{\Sigma(i,j|S)}=\Omega\left( J_{\min}\right),\quad \forall\,p\in \Nbb,
\end{equation}
from \eqref{eqn:cmibound2}. Since $\xi_{n,p}= o(J_{\min})$, \eqref{eqn:firsterm1} is $o(1)$ when $n >L\log p/J^2_{\min} $, for sufficiently large $L$ (depending on $\eta$ and $M$). For the second term in \eqref{eqn:twoerrors},
\beq \label{eqn:secondterm}
\Pbb \left[\bigcup_{(i,j) \notin G_p} \calF_1(i,j; \{\bx^n\}  ,G_p) \right] =  O \left(p^{\eta+2}\exp\left[-\frac{n
  (\xi_{n,p
 } - C_{\max}(p))^2 }{K^2}\right]\right),
\eeq
  where
\begin{equation}
 C_{\max}(p) :=  \max_{\substack{(i,j)\notin G_p}} \abs{ \Sigma(i,j| S)} =O\left(\frac{\alpha^{\gamma}}{D_{\min}}\right),\label{eqn:Imax}
\end{equation}
from \eqref{eqn:cmibound}. For the choice of $\xi_{n,p}$ in \eqref{eqn:xi},  \eqref{eqn:secondterm} is $o(1)$  and this completes the proof of Theorem~\ref{thm:corr_thres_cond}.
\qed

\subsection{Conditional Mutual Information Thresholding Test}\label{sec:cmi}

We now analyze the performance of conditional mutual information threshold test.
We first note bounds on conditional mutual information.

\begin{proposition}[Conditional Mutual Information]Under the   assumptions (A1)--(A5), we have that the conditional mutual information among non-neighbors, conditioned on the $\gamma$-local separation satisfies \beq \label{eqn:cmibound} \max_{(i,j)\notin G} I(X_i;X_j|\bfX_{S_\gamma})= O(\alpha^{2\gamma}),\eeq and the conditional mutual information among the neighbors satisfy \beq \label{eqn:cmibound2} \min_{\substack{(i,j)\in G\\ S\subset V\setminus \{i,j\}}} I(X_i;X_j|\bfX_{S})=\Omega(J^2_{\min}).\eeq \end{proposition}

 \bprf
The conditional mutual information for Gaussian variables is given by \beq\label{eqn:cmidef} I(X_i;X_j|\bfX_S)=- \frac{1}{2}\log \left[1- \rho^2(i,j|S)\right], \eeq where $\rho(i,j|S)$ is the conditional correlation coefficient, given by \[ \rho(i,j|S) :=\frac{\Sigma(i, j|S)}{\sqrt{\Sigma(i,i|S)\Sigma(j,j|S)}}. \] From \eqref{eqn:linfvar} in Proposition~\ref{prop:condcov}, we have $\Sigma(i,i|S) =O(1)$ and thus, the result holds.
\eprf

We now note the concentration bounds on empirical mutual information.

\bl[Concentration of Empirical Mutual Information]For any $p$-dimensional Gaussian random vector $\bfX=[X_1,\ldots, X_p]$,  the empirical covariance obtained from $n$ samples satisfies
\begin{equation}\label{eqn:miconc}
P( |\hI(X_i;X_j) - I(X_i;X_j)|>\epsilon) \le  24 \exp\left( -\frac{n M \epsilon^2}{ 204800  L^2  } \right),
\end{equation}for some constant $L$ which is finite when $\rho_{\max}:=\max_{i\neq j} |\rho(i,j)|<1$, and all $\epsilon<\rho_{\max}$, and for $M:=\max_{i}\Sigma(i,i)$.\el

\bprf The result on empirical covariances can be found in~\citep[Lemma 1]{Ravikumar&etal:08Arxiv}. The result in \eqref{eqn:miconc}  will be shown through a sequence of transformations. First, we will bound  $P(| \hrho(i,j) -\rho(i,j)|>\epsilon)$. Consider,
\begin{align}\nn
&P(| \hrho(i,j) -\rho(i,j)|>\epsilon)\\ \nn &= P \left(\left| \frac{{\hSigma}(i,j) }{({\hSigma}(i,i){\hSigma}(j,j))^{1/2}  } -\frac{{\Sigma}(i,j) }{({\Sigma}(i,i){\Sigma}(j,j))^{1/2}  } \right|>\epsilon \right) \\ \nn
&= P \left(\left| \frac{{\hSigma}(i,j) }{\Sigma(i,j)} \left(\frac{\Sigma(i,i)}{\hSigma(i,i) }  \frac{\Sigma(j,j)}{{\hSigma}(j,j) } \right)^{1/2}-1  \right|>\frac{\epsilon}{|\rho(i,j)|}\right) \\ \nn
&\lea P \left( \frac{{\hSigma}(i,j) }{\Sigma(i,j)}   >   \left( 1+\frac{\epsilon}{|\rho(i,j)| } \right)^{1/3} \right) +  P \left( \frac{{\hSigma}(i,j) }{\Sigma(i,j)}  < \left(  1-\frac{\epsilon}{|\rho(i,j)| }  \right)^{1/3} \right) + \ldots \nn\\
&\qquad +P \left( \frac{\Sigma(i,i)}{{\hSigma}(i,i) }  > \left(1+\frac{\epsilon}{|\rho(i,j)| }\right)^{2/3} \right) +  P \left( \frac{\Sigma(i,i)}{{\hSigma}(i,i) }  < \left(1-\frac{\epsilon}{|\rho(i,j)| }\right)^{2/3} \right)   +\ldots\nn\\
&\qquad +P \left( \frac{\Sigma(j,j)}{{\hSigma}(j,j) }  > \left(1+\frac{\epsilon}{|\rho(i,j)| }\right)^{2/3} \right) +  P \left( \frac{\Sigma(j,j)}{{\hSigma}(j,j) }  < \left(1-\frac{\epsilon}{|\rho(i,j)| }\right)^{2/3} \right) \nn  \\ \nn
&\leb  P \left( \frac{{\hSigma}(i,j) }{\Sigma(i,j)}   >   1+\frac{\epsilon}{8|\rho(i,j)| }   \right) +  P \left( \frac{{\hSigma}(i,j) }{\Sigma(i,j)}  <   1-\frac{\epsilon}{8|\rho(i,j)| }    \right) + \ldots \nn\\
&\qquad +P \left( \frac{\Sigma(i,i)}{{\hSigma}(i,i) }  >  1+\frac{\epsilon}{3|\rho(i,j)| }  \right) +  P \left( \frac{{\hSigma}(i,i) }{\Sigma(i,i)}  <  1-\frac{\epsilon}{3|\rho(i,j)|}   \right) + \ldots \nn\\
&\qquad +P \left( \frac{{\hSigma}(j,j) }{\Sigma(j,j)}  >  1+\frac{\epsilon}{3|\rho(i,j)| }  \right) +  P \left( \frac{{\hSigma}(j,j) }{\Sigma(j,j)}  <  1-\frac{\epsilon}{3|\rho(i,j)|}  \right) \nn  \\ \nn
&\lec 24 \exp\left( -\frac{n M \epsilon^2}{ 204800 |\rho(i,j)|^2   } \right)\led  24 \exp\left( -\frac{n M \epsilon^2}{ 204800  } \right)
\end{align}
where in $(a)$, we used the fact that $P(ABC >1+\delta)\le P(A> (1+\delta)^{1/3} \,\, \mathrm{ or }  \,\, B> (1+\delta)^{1/3} \,\, \mathrm{ or } \,\, C> (1+\delta)^{1/3})$ and the union bound, in $(b)$ we used the fact that $(1+\delta)^3\le 1+8\delta$ and $(1+\delta)^{-2/3} \le 1-\delta/3 $ for $\delta= \epsilon/|\rho(i,j)|<1$. Finally, in $(c)$, we used the result in \eqref{eqn:cov_conc} and in $(d)$, we used the bounds on $\rho<1$.

Now, define the bijective function $I(|\rho|) := -1/2 \log (1-\rho^2)$. Then we claim  that there exists a constant $L\in (0,\infty)$, depending only on  $\rho_{\max}<1$, such that
\begin{equation}
|I (x) - I (y)|\le L|x-y|, \label{eqn:lipsc}
\end{equation}
i.e., the function $I: [0,\rho_{\max}]\to\R^+$ is $L=L(\rho_{\max})$-Lipschitz.  This is because the slope of the function $I$ is bounded in the interval $[0,\rho_{\max}]$.    Thus, we have the inclusion
\begin{equation}
\{|\hI(X_i;X_j) - I(X_i;X_j)|>\epsilon\} \subset \{ | \hrho(i,j) -\rho(i,j)|>\epsilon /L\} \label{eqn:inclusion}
\end{equation}
since if $|\hI(X_i;X_j) - I(X_i;X_j)|>\epsilon$ it is true that $L | {\hrho}(i,j) -\rho(i,j)|>\epsilon$ from \eqref{eqn:lipsc}. We have by monotonicity of measure and \eqref{eqn:inclusion}  the desired result.
\eprf

We can now obtain the desired result on concentration of empirical conditional mutual information.

\bl[Concentration of Empirical Conditional Mutual Information]\label{lemma:cmiconc}For a walk-summable $p$-dimensional Gaussian random vector $\bfX=[X_1,\ldots, X_p]$,  we have
\begin{equation}\label{eqn:cmiconc}
P\left[ \max_{\substack{i\neq j\\ S\subset V\setminus\{i,j\}, |S|\leq \eta}} |\hI(X_i;X_j|\bfX_S) - I(X_i;X_j|\bfX_S)|>\epsilon\right] \le  24 p^{\eta+2} \exp\left( -\frac{n M \epsilon^2}{ 204800  L^2  } \right),
\end{equation}for   constants $M , L\in (0,\infty)$ and all $\epsilon<\rho_{\max}$,   where $\rho_{\max}:=\max_{\substack{i\neq j\\ S\subset V\setminus\{i,j\}, |S|\leq \eta}} |\rho(i,j|S)|$.\el

\bprf
Since the model is walk-summable, we have that $\max_{i, S} \Sigma(i,i|S) = O(1)$ and thus, the constant $M$ is bounded. Similarly, due to  strict positive-definiteness we have $\rho_{\max}<1$ even as $p\to \infty$, and thus, the constant $L$ is also finite. The result then follows from union bound. \eprf

The sample complexity for structural consistency of $\thresmialgo$ follows on lines of analysis for $\threscondalgo$.

\section{Necessary Conditions  for Model Selection}  \subsection{Necessary Conditions for Exact Recovery}\label{prf:nec}
\def\bR{\mathbf{R}}

We provide the proof of Theorem~\ref{thm:converse_gaussian} in this section.   We collect  four  auxiliary lemmata whose proofs (together with the proof of Lemma~\ref{lem:typ})  will be provided at the end of the section.  For information-theoretic notation, the reader is referred to~ \citet{Cover&Thomas:book}.

%In this section, we delineate the proof of Theorem~\ref{thm:converse_gaussian}.  The proof of Theorem~\ref{thm:converse_discrete} largely goes through fine except in two crucial places, namely the   bounds in \eqref{eqn:cardX} and \eqref{eqn:ent_pos}.
%
%Firstly, the differential entropy (see Chapter 8 of \cite{Cover&Thomas:book}) of $\bX^n$ is no longer upper bounded by $pn \log_2 |\calX|$ because  $\bX$ is a continuous random vector (with respect to the Lebesgue measure in $\bbR^p$). In other words $|\calX|$ is infinite (in fact uncountable). Secondly, the differential entropy in \eqref{eqn:ent_pos_0}, now written as $h(\bX^n|G,W)$, is no longer guaranteed to be non-negative because the differential entropy of a continuous random vector may be negative.   Hence, the lower bound in   \eqref{eqn:ent_pos}  is no longer valid.
%
%To resolve these technical issues in the proof of the Gaussian case, we employ the following two auxiliary lemmata.

\begin{lemma}[Upper Bound on Differential Entropy of Mixture] \label{lem:diff_ent}
Let $\alpha<1$. Suppose asymptotically almost surely each precision matrix $\bJ_G = \bI-\bR_G$    satisfies \eqref{eqn:walksummable}, i.e., that $\|\overline{\bR}_G\|\leq\alpha$ for a.e. $G\in \Gmsc(p)$.  Then, for the Gaussian model, we have
\begin{equation}
h(\bX^n)\le \frac{pn }{2} \log_2  \left(\frac{2\pi e }{ 1-\alpha} \right),
\end{equation}
where recall that $\bX^n|G \sim \prod_{i=1}^n f(\bx_i|G)$.
\end{lemma}

%We remark the smaller the $\delta$ in Lemma~\ref{lem:diff_ent}, the ``harder'' it is to learn the model. Thus, the value of $\delta$ provides an indication of the difficulty of the learning task.

For the sake of convenience, we define the random variable:
\begin{equation}
W = \left\{ \begin{array}{cc}
1 & G\in\calT_{\epsilon}^{(p)} \\
0 & G\notin\calT_{\epsilon}^{(p)}
\end{array} .
\right.
\end{equation}
The random variable $W$ indicates whether $G\in \calT_{\epsilon}^{(p)}$.

\begin{lemma}[Lower Bound on Conditional Differential Entropy] \label{lem:non-neg}
Suppose that each  precision matrix $\bJ_G$ has unit diagonal.  Then,
\begin{equation}
h(\bX^n |G, W) \ge -   \frac{pn}{2}\log_2(2\pi e).
\end{equation}
\end{lemma}

\begin{lemma}[Conditional Fano Inequality]\label{lem:cond_fano}
In the above notation, we have
\begin{equation}
\frac{H(G|\bX^n, G\in\calT_{\epsilon}^{(p)})-1}{\log_2(|\calT_{\epsilon}^{(p)}| -1)}\le P (\hG(\bX^n)\ne G|G\in\calT_{\epsilon}^{(p)}) .
\end{equation}
\end{lemma}

\begin{lemma}[Exponential Decay in Probability of Atypical Set]\label{lem:chernoff}
Define the rate function $K(c,\epsilon):= \frac{c}{2} [ (1+\epsilon) \ln (1+\epsilon) -\epsilon ]$. The probability of the $\epsilon$-atypical set decays as
\begin{equation}
P((\calT_{\epsilon}^{(p)} )^c) =  P(G\notin\calT_{\epsilon}^{(p)}) \le 2\exp \left( -p K(c,\epsilon) \right)   \label{eqn:chernoff}
\end{equation}
for all $p\ge 1$.
\end{lemma}
Note the non-asymptotic nature of the bound in \eqref{eqn:chernoff}. The rate function $K(c,\epsilon)$ satisfies $\lim_{\epsilon\downarrow 0} K(c,\epsilon)/ \epsilon^2 =c/4$. We prove Theorem~\ref{thm:converse_gaussian} using  these lemmata.

\bprf Consider the following sequence of lower bounds:
\begin{align}
\frac{pn}{2}\log_2\left(\frac{2\pi e}{1-\alpha}\right) &\gea h( \bX^n) \nn\\
&\geb h(\bX^n|W) \\
 &= I(\bX^n; G|W) + h(\bX^n|G,W)  \nn \\
&\gec  I(\bX^n; G|W)-   \frac{pn}{2}\log_2(2\pi e) \nn \\
&= H(G|W) - H(G |\bX^n,  W) -\frac{pn}{2}\log_2(2\pi e)  , \label{eqn:diff}
\end{align}
where $(a)$ follows from Lemma~\ref{lem:diff_ent}, $(b)$ is because conditioning does not increase differential entropy and $(c)$ follows from Lemma~\ref{lem:non-neg}. We will lower bound the first term in \eqref{eqn:diff} and upper bound the second term  in \eqref{eqn:diff}. Now consider the first term in   \eqref{eqn:diff}:
\begin{align}
H(G|W) &= H(G|W=1) P(W =1) + H(G|W=0) P(W =0) \nn \\
&\gea H(G|W=1) P(W =1) \nn\\
&\geb H(G| G\in\calT_{\epsilon}^{(p)} ) (1-\epsilon)   \nn\\
&\gec  (1-\epsilon) \binom{p}{2} \Hb \left(\frac{c}{p}\right) , \label{eqn:lowercard}
\end{align}
where $(a)$ is because the entropy  $H(G|W=0)$ and the probability $ P(W =0)$ are both non-negative. Inequality $(b)$ follows for all $p$ sufficiently large from the definition of $W$ as well as Lemma~\ref{lem:typ} part 1.  Statement $(c)$ comes from  fact that
\begin{align}\nn
H(G| G\in\calT_{\epsilon}^{(p)} ) &= -\sum_{g\in\calT_{\epsilon}^{(p)}} P(g|g\in\calT_{\epsilon}^{(p)})\log_2 P(g|g\in\calT_{\epsilon}^{(p)}) \\ \nn
& \ge   - \sum_{g\in\calT_{\epsilon}^{(p)}}  P(g|g\in\calT_{\epsilon}^{(p)})  \left[ -\binom{p}{2}\Hb\left(\frac{c}{p}\right) \right] = \binom{p}{2}\Hb\left(\frac{c}{p}\right) .
\end{align}
We are now done bounding the first term in the difference in \eqref{eqn:diff}.

Now we will  bound the second term in  \eqref{eqn:diff}. First we will derive a bound on  $H(G|\bX^n, W=1)$. Consider,
\begin{align}\nn
\Pep&:= P (\hG(\bX^n)\ne G) \\
\nn&\eqa P (\hG(\bX^n)\ne G|W=1) P(W=1)+P (\hG(\bX^n)\ne G|W=0) P(W=0)\\
\nn&\ge P (\hG(\bX^n)\ne G|W=1) P(W=1) \\
\nn&\geb P (\hG(\bX^n)\ne G|G\in\calT_{\epsilon}^{(p)})  \left(\frac{1}{1+\epsilon} \right) \\
&\gec \frac{H(G|\bX^n, G\in\calT_{\epsilon}^{(p)})-1}{\log_2 |\calT_{\epsilon}^{(p)}|}\left(\frac{1}{1+\epsilon} \right), \label{eqn:second_int}
\end{align}
where $(a)$ is by the law of total probability,  $(b)$ holds for all $p$ sufficiently large by   Lemma~\ref{lem:typ} part 1  and $(c)$ is due to the conditional version of Fano's inequality (Lemma~\ref{lem:cond_fano}).  Then, from~\eqref{eqn:second_int}, we have
\begin{align}
\nn H(G|\bX^n, W=1) &\le \Pep (1+\epsilon) \log_2 |\calT_{\epsilon}^{(p)}|  +1\\*
&\le \Pep (1+\epsilon) \binom{p}{2} \Hb\left( \frac{c}{p} \right)+1  . \label{eqn:lowercard_int}
\end{align}
Define the {\em rate function} $K(c,\epsilon):= \frac{c}{2} [ (1+\epsilon) \ln (1+\epsilon) -\epsilon ]$.    Note that this function is positive whenever $c, \epsilon>0$.   In fact it is monotonically increasing in both parameters. Now we utilize \eqref{eqn:lowercard_int} to bound $H(G|\bX^n, W)$:
\begin{align}\nn
H(G|\bX^n, W)&= H(G|\bX^n, W=1) P(W=1) + H(G|\bX^n, W=0) P(W=0) \\ \nn
&\lea H(G|\bX^n, W=1)  + H(G|\bX^n, W=0) P(W=0) \\ \nn
&\leb H(G|\bX^n, W=1)   + H(G|\bX^n, W=0) (2  e^{-p K(c,\epsilon) }) \\ \nn
&\lec H(G|\bX^n, W=1)   +  p^2  (2  e^{-p K(c,\epsilon) })\\ &\led  \Pep (1+\epsilon) \binom{p}{2} \Hb\left( \frac{c}{p} \right) +1 +  2 p^2   e^{-p  K(c,\epsilon)} , \label{eqn:lowercard2}
\end{align}
where $(a)$ is because we upper  bounded $P(W=1)$ by unity, $(b)$ follows by  Lemma~\ref{lem:chernoff}, $(c)$ follows by upper bounding the conditional entropy by $p^2$   and $(d)$ follows from \eqref{eqn:lowercard_int}.

Substituting \eqref{eqn:lowercard} and \eqref{eqn:lowercard2} back into  \eqref{eqn:diff} yields
\begin{align}\nn
\frac{pn}{2}\log_2 \left[2\pi e\left(  \frac{1}{1-\alpha} + 1\right) \right] &\ge (1-\epsilon) \binom{p}{2} \Hb \left(\frac{c}{p}\right)-  \Pep (1+\epsilon) \binom{p}{2} \Hb\left( \frac{c}{p} \right) -1 -   2 p^2   e^{-p  K(c,\epsilon)} \\ \nn
 &= \binom{p}{2} \Hb \left(\frac{c}{p}\right)  \left[ (1-\epsilon) - \Pep (1+\epsilon)\right] - \Theta(p^2 e^{-p  K(c,\epsilon) }),
\end{align}
which implies that
\begin{equation}\nn
n\ge \frac{2}{p \log_2 \left[2\pi e\left(  \frac{1}{1-\alpha} + 1\right) \right]  } \binom{p}{2} \Hb \left(\frac{c}{p}\right)  \left[ (1-\epsilon) - \Pep (1+\epsilon)\right]  - \Theta(p e^{-p  K(c,\epsilon) }) .
\end{equation}
Note that $ \Theta(p e^{-p  K(c,\epsilon) })\to 0$ as $p\to\infty$ since the rate function $K(c,\epsilon)$ is positive.  If we impose that $\Pep \to 0$ as $p\to\infty$, then $n$ has to satisfy \eqref{eqn:lower_bound_gauss} by the arbitrariness of $\epsilon>0$. This completes the proof of Theorem~\ref{thm:converse_gaussian}. \eprf

\subsection{Proof of Lemma~\ref{lem:typ}} \label{prf:lem:typ}

\bprf
Part 1 follows directed from the law of large numbers. Part 2 follows from the fact that the Binomial pmf is maximized at its mean. Hence, for $G\in\calT_{\epsilon}^{(p)}$, we have
\[
P(G)\le  \left(\frac{c}{p}\right)^{c p /2} \left(1-\frac{c}{p}\right)^{\binom{p}{2} - cp/2}.
\]
We arrive at the upper bound after some rudimentary algebra. The lower bound can be proved by observing that   for $G\in\calT_{\epsilon}^{(p)}$, we have
\begin{align}\nn
P(G)&\ge  \left(\frac{c}{p}\right)^{cp(1+\epsilon) /2} \left(1-\frac{c}{p}\right)^{\binom{p}{2} - cp(1+\epsilon)/2} \\ \nn
&= \exp_2 \left[ \binom{p}{2}  (\frac{c}{p}\log_2  \frac{c}{p}) (1+\epsilon)  + [1 - c(1+\epsilon)/p] \log_2 (1-  \frac{c}{p})\right]\\ \nn
&\ge \exp_2 \left[ \binom{p}{2}  (\frac{c}{p}\log_2  \frac{c}{p}) (1+\epsilon)  +(1+\epsilon) (1-\frac{c}{p}) \log_2 (1-  \frac{c}{p})\right] .
\end{align}
The result in Part 2  follows  immediately by appealing to the symmetry of the binomial pmf about its mean. Part 3 follows by  the following chain of inequalities:
\begin{align}\nn
1=\sum_{G\in\frakG_n} P(G)& \ge\sum_{G\in\calT_{\epsilon}^{(p)}} P(G)\ge\sum_{G\in\calT_{\epsilon}^{(p)}}\exp_2\left[ - \binom{p}{2} \Hb \left(\frac{c}{p}(1+\epsilon)\right) \right]  \\ \nn
&= |\calT_{\epsilon}^{(p)}|\exp_2\left[ - \binom{p}{2} \Hb \left(\frac{c}{p}\right)(1+\epsilon) \right] .
\end{align}
This completes the proof of the upper bound on $|\calT_{\epsilon}^{(p)}|$. The lower bound follows by noting that for sufficiently large $n$, $P(\calT_{\epsilon}^{(p)})\ge 1-\epsilon$ (by Lemma~\ref{lem:typ} Part 1). Thus,
\begin{equation}\nn
1-\epsilon\le \sum_{G\in\calT_{\epsilon}^{(p)}}P(G) \le \sum_{G\in\calT_{\epsilon}^{(p)}} \exp_2\left[ - \binom{p}{2} \Hb \left(\frac{c}{p}\right) \right] = |\calT_{\epsilon}^{(p)}|\exp_2\left[ - \binom{p}{2} \Hb \left(\frac{c}{p}\right) \right].
\end{equation}
This completes the proof.
\eprf

\subsection{Proof of Lemma~\ref{lem:diff_ent}}
\bprf   Note that the  distribution of  $\bX$ (with $G$ marginalized out) is a Gaussian mixture model given by $\sum_{G\in\frakG_p} P(G)\calN(\bzero, \bJ_{G}^{-1})$. As such the covariance matrix of $\bX$ is given by
\begin{equation}
\bSigma_{\bX} =\sum_{G\in\frakG_p} P(G) \bJ_{G}^{-1}. \label{eqn:mixture}
\end{equation}
This is not immediately obvious but it is due to the zero-mean nature of each Gaussian probability density function  $\calN(\bzero, \bJ_{G}^{-1})$. Using \eqref{eqn:mixture}, we have the following chain of inequalities:
\begin{align}\nn
h(\bX^n) &\le n h(\bX) \\ \nn
&\lea \frac{n}{2}\log_2 \left( (2\pi e)^p \det(\bSigma_{\bX}) \right) \\ \nn
&= \frac{n}{2} \left[p \log_2  (2\pi e)  +  \log_2 \det(\bSigma_{\bX})  \right]\\ \nn
&\leb \frac{n}{2} \left[p\log_2  (2\pi e)  + p \log_2 \lambda_{\max}(\bSigma_{\bX})   \right]\\ \nn
&=\frac{n}{2} \left[p\log_2  (2\pi e)  + p\log_2 \lambda_{\max} \left(  \sum_{G\in\frakG_p} P(G) \bJ_{G}^{-1} \right)    \right]\\ \nn
&\lec\frac{n}{2} \left[p\log_2  (2\pi e)  + p \log_2   \left( \sum_{G\in\frakG_p}  P(G) \lambda_{\max} \left( \bJ_{G}^{-1} \right)  \right)  \right]\\ \nn
&= \frac{n}{2} \left[p\log_2  (2\pi e)  + p \log_2   \left( \sum_{G\in\frakG_p}  P(G) \frac{1}{\lambda_{\min}  (\bJ_G)} \right)  \right]\\ \nn
&\led \frac{n}{2} \left[p\log_2  (2\pi e)  + p \log_2   \left( \sum_{G\in\frakG_p}  P(G) \frac{1}{1-\alpha} \right)  \right]\\ \nn
&= \frac{pn }{2} \log_2  \left( \frac{2\pi e}{1-\alpha} \right),
%&= \frac{m}{2} \left[n\log_2  (2\pi e)  +  \log_2 \det \left( \sum_{G\in\frakG_p} P(G) \bJ_{G}^{-1} \right) \right] \\
%&\leb\frac{m}{2} \left[n\log_2  (2\pi e)  +   \log_2(\sigma^{2n})\right] \\
%&= \frac{mn}{2} \log_2  (2\pi e \sigma^2)
\end{align}
where $(a)$ uses the maximum entropy principle \citep[Chapter 13]{Cover&Thomas:book} i.e., that the Gaussian maximizes entropy subject to an  average power constraint   $(b)$  uses the fact that the determinant of $\bSigma_{\bX}$ is upper bounded by $\lambda_{\max}(\bSigma_{\bX})^n$, $(c)$ uses the convexity of $\lambda_{\max} (\fndot)$ (it equals to the operator norm $\|\fndot\|_2$ over the set of symmetric matrices, $(d)$ uses the fact that $\alpha \ge \| \overline{\bR}_G \|_2 \ge \|  \bR_G \|_2  = \| \bI - \bJ_G \|_2 = \lambda_{\max} (\bI - \bJ_G ) = 1-\lambda_{\min} (\bJ_G)
    $  a.a.s. This completes the proof.
\eprf

\subsection{Proof of Lemma~\ref{lem:non-neg}}
\bprf
Firstly, we lower bound $ h(\bX^n |G, W=1)$ as follows:
\begin{align}\nn
h(\bX^n |G) &= \sum_{g\in \frakG_p} P(g) h(\bX^n | G=g )\\ \nn
&\eqa n\sum_{g\in\frakG_p} P(g) h(\bX  | G=g) \\ \nn
&\eqb \frac{n}{2}\sum_{g\in\frakG_p} P(g)\log_2[ (2\pi e)^p \det(\bJ_g^{-1}) ]\\ \nn
&= -\frac{n}{2}\sum_{g\in\frakG_p} P(g)\log_2[ (2\pi e)^p \det(\bJ_g ) ]\\ \nn
&\gec -\frac{n}{2}\sum_{g\in \frakG_p} P(g)\log_2[ (2\pi e)^p   ] \\ \nn
&\ge  - \frac{pn }{2}\log_2(2\pi e),
\end{align}
where $(a)$ is because the samples in $\bX^n$ are conditionally independent given $G=g$, $(b)$ is by the Gaussian assumption, $(c)$ is by Hadamard's inequality \begin{equation}\det(\bJ_g)\le \prod_{i=1}^p [\bJ_g]_{ii}=1 \end{equation} and the assumption that  each diagonal element of each precision matrix $\bJ_g=\bI-\bR_g$ is equal to 1 a.a.s.   This proves the claim.
\eprf

\subsection{Proof of Lemma~\ref{lem:cond_fano}}
\bprf
Define the ``error'' random variable
\begin{equation}\nn
E = \left\{ \begin{array}{cc}
 1 &\hG(\bX^n)\ne G\\% \mbox{ and } W = 1\\
0 &\hG(\bX^n) = G %\mbox{ and } W = 1
\end{array}
\right. .
\end{equation}
Now consider
\begin{align}
H(E,G |\bX^n, W=1)  &= H(E  |\bX^n, W=1)  + H( G |E,\bX^n, W=1) \label{eqn:first_exp}   \\*
&= H(G  |\bX^n, W=1)  + H(E |G, \bX^n, W=1) . \label{eqn:sec_exp}
\end{align}
The first term in \eqref{eqn:first_exp} can be bounded above by  $1$ since the alphabet of the random variable $E$ is of size 2.  Since $H( G |E=0,\bX^n, W=1)=0$, the second term in  \eqref{eqn:first_exp} can be  bounded from above as
\begin{align}
H( G |E, \bX^n, W=1) &=H( G |E=0 , \bX^n, W=1) P(E=0| W=1) \nn\\ \nn
&\qquad\qquad+ H( G |E=1, \bX^n, W=1) P(E=1| W=1) \\ \nn
&\le P (\hG(\bX^n)\ne G|G\in\calT_{\epsilon}^{(p)}) \log_2( |\calT_{\epsilon}^{(p)}|-1) .
\end{align}
The second term in \eqref{eqn:sec_exp}  is 0. Hence, we have the desired conclusion.
\eprf

\subsection{Proof of Lemma~\ref{lem:chernoff}}
\bprf The proof uses standard Chernoff bounding techniques but the scaling in $p$ is somewhat different from the usual Chernoff (Cram\'{e}r) upper bound.
For simplicity, we will use $M := \binom{p}{2}$. Let $Y_i , i=1,\ldots, M$ be independent Bernoulli random variables such that $P(Y_i =1)= c/p$. Then the probability in question can be bounded as
\begin{align}\nn
P(G\notin\calT_{\epsilon}^{(p)})  & = P\left( \left|\frac{1}{cp} \sum_{i=1}^M Y_i -\frac{1}{2}\right| >\frac{ \epsilon}{2}\right)\\ \nn
 & \lea  2P\left( \frac{1}{cp} \sum_{i=1}^M Y_i    >  \frac{1+\epsilon}{2}\right)\\
%  & =  2\Ebb\left[ \bbI \left\{ \frac{1}{n} \sum_{i=1}^M Y_i -     (1+\epsilon)\frac{c\epsilon}{2} >0\right\}\right]\\
    & \leb  2\Ebb\left[ \exp \left( t \sum_{i=1}^M Y_i - p \, t \frac{c}{2}(1+\epsilon)  \right)\right] \label{eqn:introt} \\
&= 2\exp \left( -p \, t \frac{c}{2} (1+\epsilon) \right)     \prod_{i=1}^M \Ebb [\exp(t Y_i)] \label{eqn:chernoff_chain} ,
\end{align}
where $(a)$ follows from the union bound, $(b)$ follows from  an application of Markov's inequality with $t\ge 0$ in \eqref{eqn:introt}.  Now, the moment generating function of a Bernoulli random variable with probability of success $q$ is $qe^t + (1-q)$. Using this fact, we can further upper bound \eqref{eqn:chernoff_chain} as follows:
\begin{align}\nn
P(G\notin\calT_{\epsilon}^{(p)})  &  =2 \exp \left(- p\,  t \frac{c}{2} (1+\epsilon) + M \ln (\frac{c}{p} e^t  + (1-\frac{c}{p} ) \right) \\ \nn
&  \lea 2\exp \left(-  p\, t \frac{c}{2} (1+\epsilon) +\frac{p(p-1)}{2}  \frac{c}{p} (e^t-1)   \right) \\
&  \le  2 \exp \left(- p \left[ t \frac{c}{2} (1+\epsilon) -\frac{c}{2} (e^t-1)  \right]  \right) , \label{eqn:chernoff_chain2}
\end{align}
where in $(a)$,  we used the fact that $\ln(1+z)\le z$ . Now, we differentiate the exponent in square brackets  with respect to $t\ge 0$ to find the tightest bound. We observe that the optimal parameter is $t^* = \ln (1+\epsilon)$. Substituting this back into  \eqref{eqn:chernoff_chain2} completes the proof.
\eprf

\subsection{Necessary Conditions for  Recovery with Distortion} \label{sec:proof_distort}
We now provide the proof for  Corollary \ref{cor:converse_gauss_gen}.

\def\bR{\mathbf{R}}

The proof of Corollary~\ref{cor:converse_gauss_gen} follows from the following generalization of the conditional Fano's inequality presented in Lemma~\ref{lem:cond_fano}. This is a modified version of an analogous theorem in \citep{Kim08}.
\begin{lemma}[Conditional Fano's Inequality (Generalization)] \label{lem:gen_fano}
In the above notation, we have
\begin{equation}
\frac{H(G|\bX^n, G\in\calT_{\epsilon}^{(p)} )-1-\log_2 L}{\log_2(|\calT_{\epsilon}^{(p)}| -1)}\le P (d(G,\hG(\bX^n)) > D|G\in\calT_{\epsilon}^{(p)}) \label{eqn:gen_fanos}
\end{equation}
where $L= \binom{p}{2}\Hb(\beta)$ and $\beta$ is defined in \eqref{eqn:alpha_def}.
\end{lemma}
We will only provide a proof sketch of Lemma~\ref{lem:gen_fano} since it is similar to Lemma~\ref{lem:cond_fano}.
\bprf
The key to establishing \eqref{eqn:gen_fanos} is to upper bound the cardinality of the set $\{G\in\frakG_p : d(G, G') \le D\}$, which is isomorphic to $\{E \in \frakE_p: |E\triangle E'|\le D\}$, where $\frakE_p$ is the set of all edge sets (with $p$ nodes). For this purpose, we order the node pairs in a labelled undirected graph lexicographically. Now, we map each edge set $E$ into a length-$\binom{p}{2}$ bit-string $s(E) \in \{0,1\}^{\binom{p}{2}}$. The characters in the string $s(E)$ indicate whether or not an edge is present between two node pairs. Define $d_H(s, s')$ to be the Hamming distance between strings $s$ and $s'$. Then, note that
\begin{equation}
|E\triangle E'| = d_H(s(E), s(E')) = d_H  (s(E) \oplus   s(E'), 0  ) \label{eqn:hamming}
\end{equation}
where $\oplus$ denotes addition in $\Fbb_2$ and $0$ denotes the all zeros string. The relation in \eqref{eqn:hamming} means that  the cardinality of the set $\{E  \in \frakE_n:  |E\triangle E'|\le D\}$ is equal to the number of strings of Hamming weight less than or equal to $D$. With this realization, it is easy to see that
\begin{equation}
|\{s \in \{0,1\}^{\binom{p}{2}}: d_H(s, 0)\le D\}| = \sum_{k=1}^D\binom{\binom{p}{2} }{k} \le 2^{ \binom{p}{2}  \Hb (D/\binom{p}{2} ) }=2^L.\nn
\end{equation}
By using the same steps as in the proof of  Lemma~\ref{lem:gen_fano} (or Fano's inequality for list decoding), we arrive at the desired conclusion.
\eprf 

\end{appendix}
{

%\bibliographystyle{IEEEtran}
% \bibliography{\bibhome/Book,\bibhome/Journal,\bibhome/Conf, \bibhome/Misc,../isitbib}
%
}

\end{document}